\documentclass[10pt,twocolumn,letterpaper]{article}

\usepackage[pagenumbers]{cvpr} 

\usepackage[dvipsnames]{xcolor}

\definecolor{cvprblue}{rgb}{0.21,0.49,0.74}
\usepackage[pagebackref,breaklinks,colorlinks,citecolor=cvprblue]{hyperref}
\usepackage{algorithm,algorithmic}
\usepackage{multirow}
\usepackage{amsmath}
\usepackage[linewidth=1pt]{mdframed}
\usepackage{tabularx}
\usepackage[accsupp]{axessibility} 

\def\paperID{****} 
\def\confName{CVPR}
\def\confYear{2024}

\title{Simple Semantic-Aided Few-Shot Learning}

\author{Hai Zhang$^{1}$\footnotemark[1]\quad
Junzhe Xu$^{1,2}$\footnotemark[1]\hspace{0.15cm}\footnotemark[3]\quad
Shanlin Jiang$^{3}$\quad
Zhenan He$^{1}$\footnotemark[2]\quad \\
\textsuperscript{1}College of Computer Science, Sichuan University \\
\textsuperscript{2}Banma Network Technology, Alibaba Group\\
\textsuperscript{3}Naveen Jindal School of Management, University of Texas at Dallas\\
{\tt\small zhanghi@stu.scu.edu.cn, xujunzhe.xjz@alibaba-inc.com}\\
{\tt\small shanlin.jiang@utdallas.edu, zhenan@scu.edu.cn}
}

\begin{document}
\maketitle

\renewcommand{\thefootnote}{\fnsymbol{footnote}} 
\footnotetext[1]{Equal Contribution. $^{\dag}$\text{Corresponding Author.}} 
\footnotetext[3]{Work done during master's degree at Sichuan University.} 
\begin{abstract}
   Learning from a limited amount of data, namely Few-Shot Learning, stands out as a challenging computer vision task. Several works exploit semantics and design complicated semantic fusion mechanisms to compensate for rare representative features within restricted data. However, relying on naive semantics such as class names introduces biases due to their brevity, while acquiring extensive semantics from external knowledge takes a huge time and effort. This limitation severely constrains the potential of semantics in Few-Shot Learning. In this paper, we design an automatic way called Semantic Evolution to generate high-quality semantics. The incorporation of high-quality semantics alleviates the need for complex network structures and learning algorithms used in previous works. Hence, we employ a simple two-layer network termed Semantic Alignment Network to transform semantics and visual features into robust class prototypes with rich discriminative features for few-shot classification. The experimental results show our framework outperforms all previous methods on six benchmarks, demonstrating a simple network with high-quality semantics can beat intricate multi-modal modules on few-shot classification tasks. Code is available at \url{https://github.com/zhangdoudou123/SemFew}.
\end{abstract}

\section{Introduction}
Deep learning models have witnessed substantial advancements by leveraging extensive annotated data \cite{deng2009imagenet}. Nevertheless, in numerous real-world scenarios, the shortage of labeled data restricts the applicability of conventional deep learning approaches. In contrast, humans possess a remarkable cognitive ability to learn new concepts and recognize categories from just a few examples \cite{biederman1987recognition}. Motivated by this, Few-Shot Learning (FSL) \cite{snell2017prototypical} is proposed to mimic the ability of humans to learn from a few labeled samples.

In a typical setting of FSL \cite{snell2017prototypical}, the support set, which consists of $N$ novel classes and each class contains $K$ samples, is provided for the model. The model learned on the support set is required to accurately classify test samples, which are labeled as the query set, into one of the $N$ categories. A conventional classification approach involves projecting both support and query sets onto a pre-established metric space. Then, query images are classified by finding the nearest support image. However, few support samples may not contain sufficient discriminative features for recognition. These samples are typically located on the periphery of the sample cluster in the metric space, resulting in unstable distance evaluations and inaccurate classification.

To address this issue, researchers start from visual-based methods \cite{vinyals2016matching,snell2017prototypical,oreshkin2018tadam,finn2017model,li2019finding,mangla2020charting,cheng2023frequency,tian2020rethinking,ye2020few,hiller2022rethinking,dong2022self,hao2023class}, aiming at extracting class-related features from periphery image, reducing the intra-class variation among samples, and thus constructing robust features for classification. Although visual-based methods have undergone extensive investigation and achieved great success, this type of approach still struggles to handle situations when there is only one hard periphery sample with minimal semantic features per class, \ie, the one-shot learning task \cite{koch2015siamese,vinyals2016matching}. Hence, some researchers turned to applying semantics as auxiliary information to help the model better understand periphery images, leveraging the synergies between language and vision modalities.

From this perspective, semantic-based methods \cite{xing2019adaptive,peng2019few,li2020boosting,xu2022generating,zhang2021prototype,schonfeld2019generalized,wang2019tafe,tokmakov2019learning,yan2021aligning,chen2023semantic} dedicated to exploring different types of semantics to enhance periphery samples. One kind of method \cite{xing2019adaptive,yan2021aligning,xu2022generating} applies the class name as the semantic source. However, class name is not the best way to understand a novel class. For instance, If a person never saw a zebra, it would be easier to identify a zebra by the definition, \ie, a horse with stripes, than by the name zebra. Meanwhile, the class name contains ambiguity in some scenarios. For example, the class \textit{ear} is easily misidentified as the human organ, but images in MiniImageNet \cite{vinyals2016matching} belong to this class are actually photos of corn. Another kind of semantic exploits discrete attribute labels as the substitution of mere class names and achieves good performance \cite{zhang2021prototype}. Nevertheless, the collection process is time-consuming and expert assistance requiring. Moreover, discrete attribute labels lack robust representation and struggle to benefit from context. Therefore, effectively collecting and leveraging high-quality semantics becomes an urgent research need. 

\begin{figure}
    \centering
    \includegraphics[width=\linewidth]{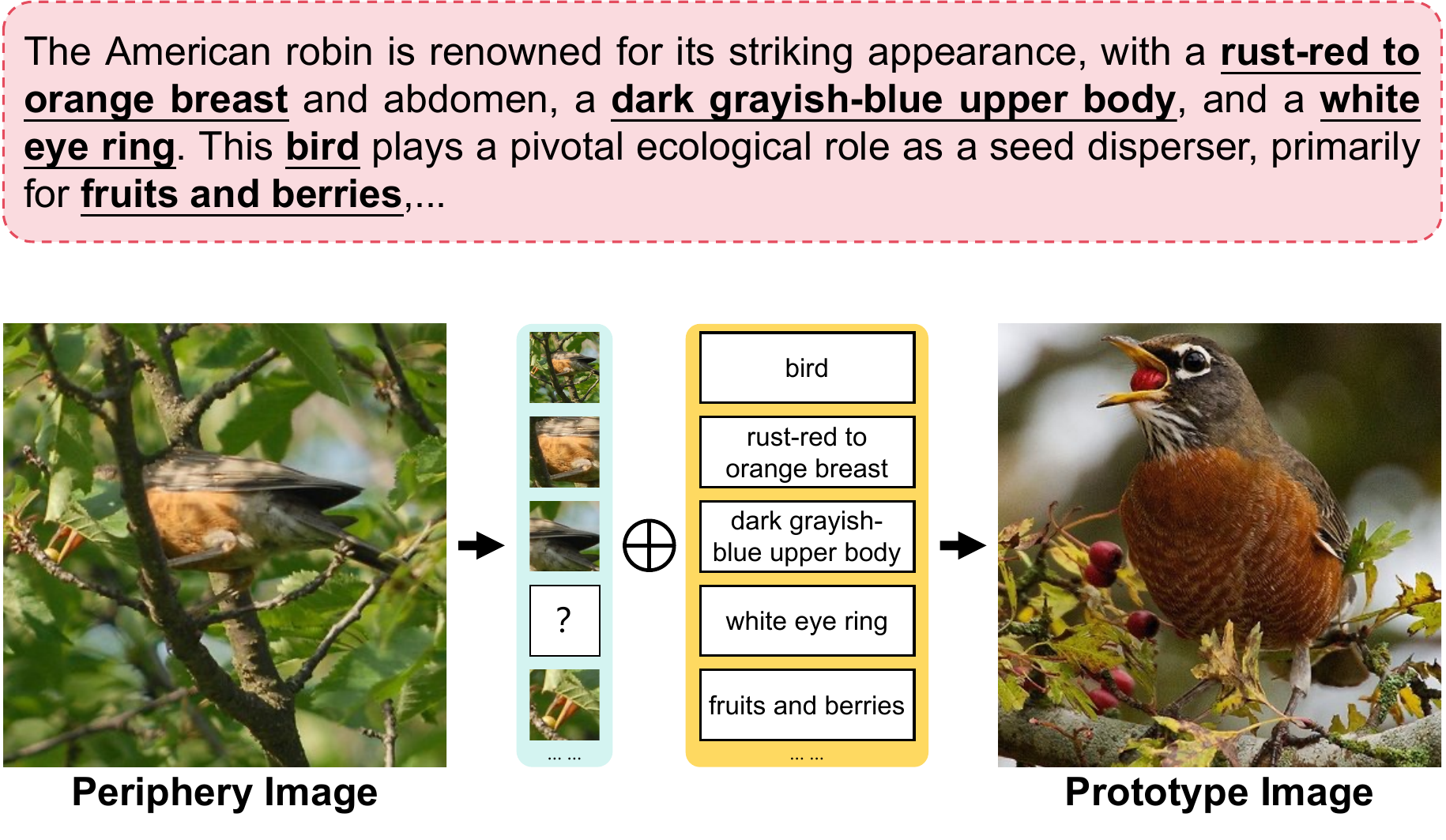}
    \caption{The introduction of how high-quality semantics reconstruct the class prototype through complementarity between modalities. Periphery image is an image with fewer discriminative features and prototype image represents images with concrete and enriched representative features.}
    \label{fig:fig1}
\end{figure}

In this paper, we propose a simple framework called \textbf{Sem}antic-Aided \textbf{Few}-Shot Learning (SemFew). To begin with, we identify the vital role of high-quality semantics in FSL and innovatively propose an automatic step-by-step \textbf{Sem}antic \textbf{Evo}lution (SemEvo) process to acquire detailed and accurate semantics. It first converts the class name into a short description that matches the image content of each class. Next, to further enrich recognizable characteristics, we expand and paraphrase short descriptions. The paraphrased descriptions contain more class-related knowledge than class names and short descriptions, making them better encompass a vast array of visual features. To illustrate, \cref{fig:fig1} showcases an instance that the paraphrased semantic contains a large number of detailed descriptions matching visual features \eg organ characteristics and eating habitat. With this information, the model can better reconstruct an image with periphery feature (left image), into an image with prototype feature (right image), which benefits few-shot classification. Then, to fully exploit the advantage of paraphrased semantics, we design a network called \textbf{Sem}antic \textbf{Align}ment Network (SemAlign). Different from previous methods, SemAlign does not adopt complex semantic understanding modules but utilizes a basic two-layer network. After accepting semantics and visual features as input, it reconstructs prototype features for subsequent classification. Experimental results on six widely adopted benchmarks demonstrate the effectiveness of our method, providing a simple insight into future FSL research. Our contributions can be summarized as three points:
\begin{itemize}
\item[$\bullet$]
To the best of our knowledge, we are the first to consider the automatic way of collecting high-quality semantics and applying them in FSL.
\item[$\bullet$] 
We design a simple and efficient way to translate high-quality semantics and visual features into prototypes, without any intricate semantic understanding modules.
\item[$\bullet$]
Our approach achieves state-of-the-art performance across six benchmarks in FSL research, underscoring that a basic network can obtain excellent performance when supported by high-quality semantics.
\end{itemize}

\section{Related Work}
The significant issue in FSL is how to effectively and accurately extract class-related features from limited data. Targeting this issue, existing research explored two different types of approaches, \ie, visual-based methods and semantic-based methods.

\textbf{Visual-Based Methods} focus on extracting class-related features from images for classification. There are two main types of this approach. The approaches with the optimization principle behind \cite{ravi2017optimization,andrychowicz2016learning,finn2017model} concentrated on acquiring a set of initial model parameters capable of rapid adaptation to novel classes. However, updating the entire model with a limited amount of labeled data may result in meta-overfitting \cite{mishra2017simple,zintgraf2019fast,rusu2018meta,jamal2019task,elsken2020meta}. The approaches considered metric design aiming at training a metric space, where inter-class distances are maximized, while intra-class distances are minimized. Within this type of approach, one attracting method focuses on integrating self-supervised loss to pretrain the backbone network \cite{mangla2020charting,doersch2020crosstransformers,gidaris2019boosting,hiller2022rethinking,tian2020rethinking}. Other methods are dedicated to improving the metric, and previous research spanning from initially manually designed metrics \cite{snell2017prototypical,koch2015siamese,zhang2022deepemd,bateni2020improved} to subsequently model-based metrics \cite{vinyals2016matching,sung2018learning}, all of which have yielded encouraging outcomes.

\textbf{Semantic-Based Methods} sought to enhance visual recognition performance by fusing complementary information from both visual and textual modalities \cite{xing2019adaptive,zhang2021prototype,xu2022generating,li2020boosting,chen2023semantic}, and related research introducing intricate network frameworks to effectively utilize textual information. Chen \etal \cite{xing2019adaptive} introduced an adaptive fusion mechanism to integrate visual prototypes with semantic prototypes obtained through word embeddings of class labels. Zhang \etal \cite{zhang2021prototype} designed a framework for prototype completion, leveraging rare attribute-level information from labels to construct representative class prototypes. Xu and Le \cite{xu2022generating} proposed the use of the Conditional Variational Autoencoder (CVAE) \cite{sohn2015learning} model to generate visual features based on semantic embeddings for further classification. Li \etal \cite{li2020boosting} presented a task-relevant adaptive margin loss based on the semantic similarity between class names. Chen \etal \cite{chen2023semantic} devised complementary mechanisms for inserting semantic prompts into the feature extractor, allowing the feature extractor to better focus on class-specific features.

Previous research emphasized the exploitation of class names and designed a complicated network to integrate semantics with visual information. In contrast, our primary concern lies in the inherent quality of semantics, and how it can relieve the burden of designing intricate network structures and learning algorithms. Remarkably, we exhibit that high-quality semantics information has the ability to yield superior outcomes even within a simple network.

\section{Methodology}
This section first revisits the background of FSL, then introduces two components of SemFew: explaining how semantics are gradually improved through SemEvo, and demonstrating the details of our proposed SemAlign.

\subsection{Background}
The dataset of FSL includes two components: a base set $\mathcal{D}_{base} = \{(x, y)|x\in \mathcal{X}_{base}, y\in \mathcal{C}_{base}\}$ for training a metric space and a novel set $\mathcal{D}_{novel} = \{(x, y)|x\in \mathcal{X}_{novel}, y\in \mathcal{C}_{novel}\}$ for testing, where $x$ denotes the image and $y$ represents the label. It is important to note that the label space for both sets is disjoint, \ie, $\mathcal{C}_{base} \cap\mathcal{C}_{novel}=\emptyset$. During the testing process, the support set $\mathcal{S} = {\{(x_i, y_i)\}}_{i=0}^{N \times K}$ is randomly sampled from $\mathcal{D}_{novel}$, which contains $N$ classes with each class consisting of $K$ samples. Then, the model is required to correctly classify images in the query set $\mathcal{Q} = {\{(x_i, y_i)\}}_{i=0}^{N\times M}$ into one of $N$ classes in the support set $\mathcal{S}$, where $M$ is the number of query samples in each class. Generally, this type of classification task is named as $N$-way $K$-shot task.

A simple but effective FSL method is to calculate the prototype of each support class, and then classify query samples by finding the nearest prototype in the metric space. To be specific, the prototype of class $t$, labeled as $p_t$, is the mean vector of all samples in the support set belonging to class $t$: 
\begin{equation} \label{eq1}
p_t=\dfrac{1}{|\mathcal{S}_t|}\sum_{\substack{x_i \in \mathcal{S}_t}} f(x_i),
\end{equation}
where $\mathcal{S}_t$ is a set of all samples belonging to class $t$ in the support set $\mathcal{S}$ and $f$ denotes the backbone network. Then, 
the probability of a query sample $q$ belonging to class $t$ is related to the distance between $q$ and prototype of class $t$ \cite{snell2017prototypical}, which is calculated as follows:
\begin{equation} 
P(y=t)=\dfrac{\exp{d(q,p_t)}}{\sum_{i=1}^{N}\exp{d(q,p_i)}},\label{con:probability}
\end{equation}
where $d(\cdot,\cdot)$ means distance metric, \eg, cosine distance, and $N$ denotes the number of classes.

\begin{figure}
    \centering
    \includegraphics[width=\linewidth]{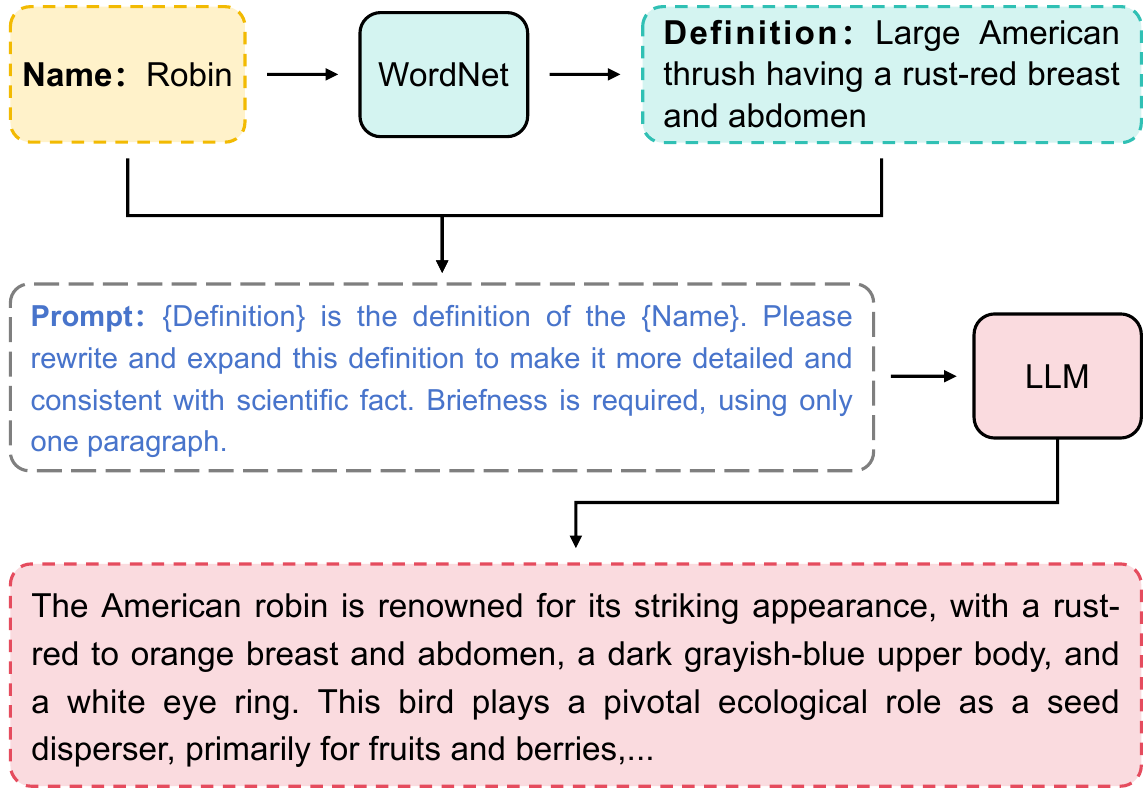}
    \caption{The illustration of how Semantic Evolution converts the simple name and the definition into the high-quality description.}
    \label{fig:fig2}
\end{figure}

\subsection{Semantic Evolution}\label{semevo}
This section explains how we automatically generate high-quality semantics for few-shot classification by the designed process termed Semantic Evolution. This process first converts class names to short descriptions and then paraphrases them into high-quality semantics.

As we illustrated before, adopting class names as semantics \cite{xu2022generating,chen2023semantic,radford2021learning} unavoidably introduces ambiguity. Therefore, we first retrieve the definition of each class name from WordNet \cite{miller1995wordnet} that matches image content, where definitions are more detailed and concrete than mere class names. However, even though definitions are able to precisely describe a class to some extent, brief descriptions may still overlook key visual information, due to the fact that visual features are highly variable.

To comprehensively conclude variable visual features, we leverage the extensive knowledge stored in the pre-trained Large Language Models (LLMs) \cite{brown2020language} to augment definitions. The prompt we used is: \textit{\{WordNet Definition\} is the definition of the \{Class Name\}. Please rewrite and expand this definition to make it more detailed and consistent with scientific fact. Briefness is required, using only one paragraph.} The illustration of the Semantic Evolution process is shown in \cref{fig:fig2}.

The paraphrased class description generated by LLMs precisely describes the corresponding class itself, which does not apply to other categories. This characteristic ensures that semantic descriptions do not leak information to new categories, which follows the paradigm of FSL.

\begin{figure*}
    \centering
    \includegraphics[width=0.95\textwidth]{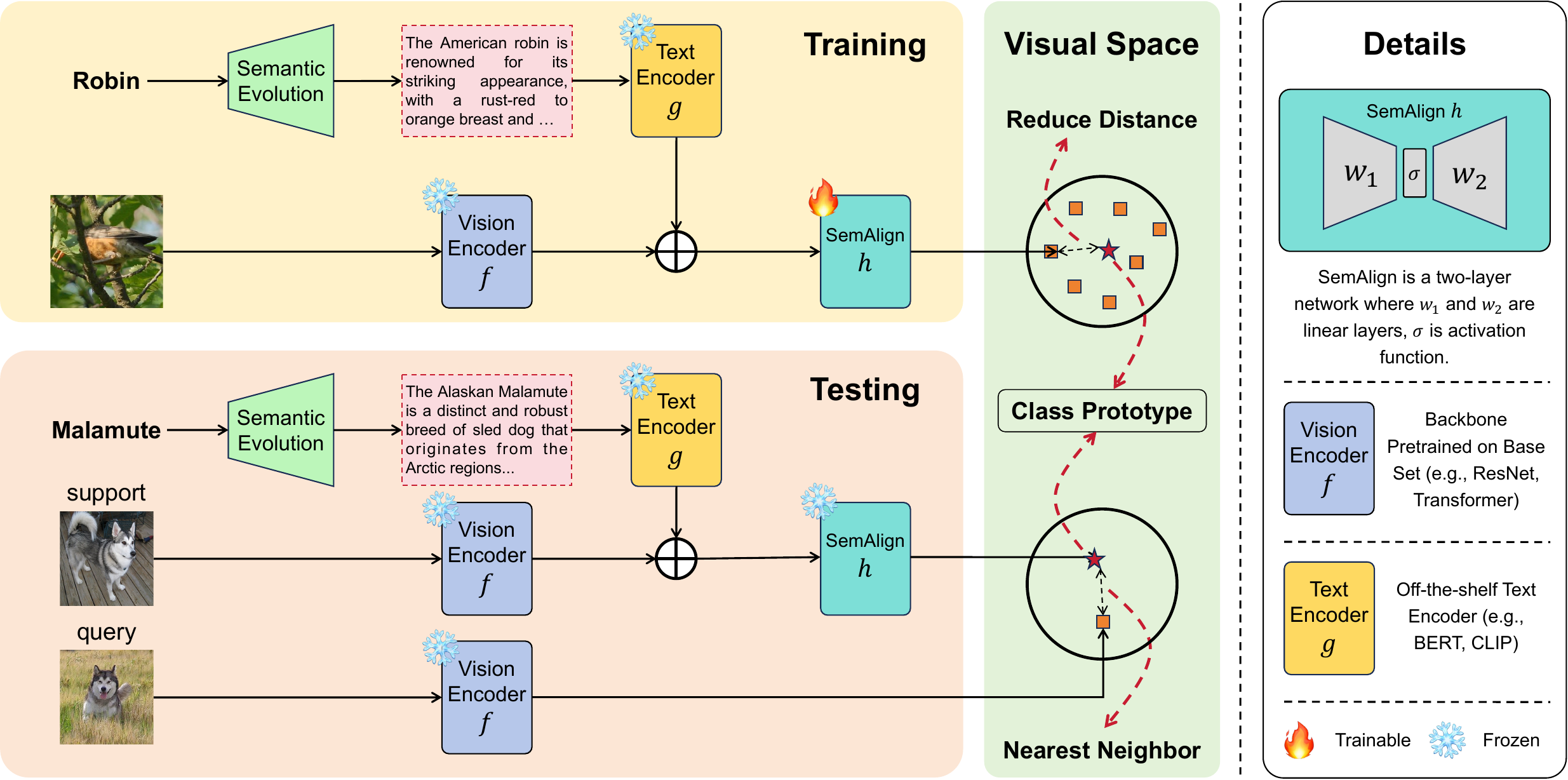}
    \caption{The framework of our proposed SemFew. During the training stage, images and paraphrased semantics are encoded and fed into SemAlign $h$, with the objective of reducing the distance between the output of $h$ and the class prototype in the visual space. During the testing stage, images in the support set are transformed into class prototypes by $h$, and query images are classified by identifying the nearest prototype. The symbol $\oplus$ denotes a concatenation operation.}
\label{fig:fig3}
\end{figure*}

\subsection{Semantic Alignment Network}\label{sec3.3:semfew}
After obtaining paraphrased semantics in the previous section, we devise the Semantic Alignment Network (SemAlign), aiming at translating hard periphery samples into robust class prototypes.

For a cluster of samples, the cluster center best encapsulates the characteristic of a category and has been regarded as a prototype in previous work \cite{snell2017prototypical}. Suppose $S$ represents semantic features, and $C$ stands for the center of each class, the simplest implementation of SemAlign is to learn the alignment between semantics $S$ and centers $C$: 
\begin{equation}
\min_{W_1,W_2}L(\sigma(S^\top W_1)W_2, C), \label{align}
\end{equation}
where $W_1$ and $W_2$ are parameters of network, $\sigma$ is an activation function.

Considering that each modality views things from a different and complementary perspective, this phenomenon can be reasonably utilized to obtain better prototypes. Therefore, we fuse visual features with high-quality semantics, and then align multi-modal output with the cluster center, the details of the whole pipeline can be found in \cref{fig:fig3}. Specifically, suppose the $i$-th image in the dataset is represented as $x_i$, its label and semantic are denoted as $y_i$ and $s_i$, respectively. First, $x_i$ is encoded by a vision encoder $f$, which is pre-trained on the base set. Also, semantic description $s_i$ is encoded by an off-the-shelf text encoder $g$. Then, the encoded visual and text features are concatenated together and fed into SemAlign $h$ to reconstruct the prototype. The training loss supervises the distance between the reconstructed prototype towards the cluster center:
\begin{equation}
    \min_{W_1,W_2}\mathbb{E}[||h(f(x_i), g(s_i)) - c_y||_1],
\end{equation}
where $h(f(x_i), g(s_i)) = \sigma([f(x_i)\cdot g(s_i)]^\top W_1)W_2$ is SemAlign with learnable weights $W_1$ and $W_2$, $[\cdot]$ means concatenation operation. 

During the testing process, a $N$-way $K$-shot support set is randomly sampled in the novel set. For class $t$, we obtain its paraphrased description $s_t$ by Semantic Evolution proposed in \cref{semevo}. Then, support image $x$ and description $s_t$ are encoded by $f$ and $g$, respectively, and fed into $h$ to get the reconstructed prototype:
\begin{equation}
r_t = \frac{1}{K}\sum_{i=1}^K h(f(x_i), g(s_t)),
\end{equation}
where $r_t$ is the reconstructed prototype. To ensure the reconstructed prototype is further aligned with the ground-truth class prototype, it will be fused with support images in a convex combination manner \cite{xing2019adaptive}:
\begin{equation}
p_t = k r_t + (1 - k) u_t,\label{con:convex}
\end{equation}
where $p_t$ means the classification prototype of class $t$, $u_t=\frac{1}{K}\sum_{i=1}^K f(x_i)$ is the mean vector of support samples, $r_t$ is the reconstructed prototype produced by SemAlign, and $k\in [0,1]$ is a manually controlled fusion factor. Then, \cref{con:probability} is utilized to assign label for each query sample, where $d(\cdot,\cdot)$ is set as cosine distance.
\begin{table*}[ht!]
    \centering
    \begin{tabular}{c | l c c c c c c}
    \toprule
         &\multirow{2}{*}{\textbf{Method}} & \multirow{2}{*}{\textbf{Venue}} & \multirow{2}{*}{\textbf{Backbone}} & \multicolumn{2}{c}{\textbf{MiniImageNet}} & \multicolumn{2}{c}{\textbf{TieredImageNet}}\\ 
        &\multirow{2}{*}{} & \multirow{2}{*}{} & \multirow{2}{*}{} & \textbf{5-way 1-shot} & \textbf{5-way 5-shot}  & \textbf{5-way 1-shot} & \textbf{5-way 5-shot}\\ 
        \midrule
        \multirow{12}{*}{\rotatebox{90}{\textbf{Visual-Based}}}
        &MatchNet \cite{vinyals2016matching} & NeurIPS'16 & ResNet-12 & 65.64 ± 0.20 & 78.72 ± 0.15 & 68.50 ± 0.92 & 80.60 ± 0.71 \\
        &ProtoNet \cite{snell2017prototypical} & NeurIPS'17 & ResNet-12 & 62.39 ± 0.21 & 80.53 ± 0.14 & 68.23 ± 0.23 & 84.03 ± 0.16 \\
        &MAML \cite{finn2017model} & ICML'17 & ResNet-12 & 49.24 ± 0.21 & 58.05 ± 0.10 & 67.92 ± 0.17 & 72.41 ± 0.20 \\
        &TADAM \cite{oreshkin2018tadam} & NeurIPS'18 & ResNet-12 & 58.50 ± 0.30 & 76.70 ± 0.30 & 62.13 ± 0.31 & 81.92 ± 0.30 \\
        &CAN \cite{hou2019cross} & NeurIPS'19 & ResNet-12 & 63.85 ± 0.48 & 79.44 ± 0.34 & 69.89 ± 0.51 & 84.23 ± 0.37 \\
        &CTM \cite{li2019finding} & CVPR'19 & ResNet-18 & 64.12 ± 0.82 & 80.51 ± 0.13 & 68.41 ± 0.39 & 84.28 ± 1.73 \\
        &RFS \cite{tian2020rethinking} & ECCV'20 & ResNet-12 & 62.02 ± 0.63 & 79.64 ± 0.44 & 69.74 ± 0.72 & 84.41 ± 0.55 \\
        &FEAT \cite{ye2020few} & CVPR'20 & ResNet-12 & 66.78 ± 0.20 & 82.05 ± 0.14 & 70.80 ± 0.23 & 84.79 ± 0.16\\
        &Meta-Baseline \cite{chen2021meta} & ICCV'21 & ResNet-12 & 63.17 ± 0.23 & 79.26 ± 0.17 & 68.62 ± 0.27 & 83.29 ± 0.18 \\
        &SUN \cite{dong2022self} & ECCV'22 & ViT-S & 67.80 ± 0.45 & 83.25 ± 0.30 & 72.99 ± 0.50 & 86.74 ± 0.33\\
        &FewTURE \cite{hiller2022rethinking} & NeurIPS'22 & Swin-T & 72.40 ± 0.78 & 86.38 ± 0.49 & 76.32 ± 0.87 & \textbf{89.96 ± 0.55} \\
        &FGFL \cite{cheng2023frequency} & ICCV'23 & ResNet-12 & 69.14 ± 0.80 & 86.01 ± 0.62 & 73.21 ± 0.88 & 87.21 ± 0.61 \\
        &Meta-AdaM \cite{metaAdaM} & NeurIPS'23 & ResNet-12 & 59.89 ± 0.49 & 77.92 ± 0.43 & 65.31 ± 0.48 & 85.24 ± 0.35 \\
        \midrule
        \multirow{9}{*}{\rotatebox{90}{\textbf{Semantic-Based}}}
        &KTN \cite{peng2019few} & ICCV'19 & Conv-128 & 64.42 ± 0.72 & 74.16 ± 0.56 & 74.16 ± 0.56 & - \\
        &AM3 \cite{xing2019adaptive} & NeurIPS'19 & ResNet-12 & 65.30 ± 0.49 & 78.10 ± 0.36 & 69.08 ± 0.47 & 82.58 ± 0.31 \\
        &TRAML\cite{li2020boosting}  & CVPR'20 & ResNet-12 & 67.10 ± 0.52 & 79.54 ± 0.60 & - & -\\
        &AM3-BERT \cite{yan2021aligning} & ICMR'21 & ResNet-12 & 68.42 ± 0.51 & 81.29 ± 0.59 & 77.03 ± 0.85 & 87.20 ± 0.70 \\
        &SVAE-Proto \cite{xu2022generating} & CVPR'22 & ResNet-12 & 74.84 ± 0.23 & 83.28 ± 0.40 & 76.98 ± 0.65 & 85.77 ± 0.50\\
        &SP-CLIP \cite{chen2023semantic} & CVPR'23 & Visformer-T & 72.31 ± 0.40 & 83.42 ± 0.30 & 78.03 ± 0.46 & 88.55 ± 0.32 \\ 
        \cmidrule(lr){2-8}
        &SemFew & Ours & ResNet-12 & 77.63 ± 0.63 & 83.04 ± 0.48 & 78.96 ± 0.80 & 85.88 ± 0.62 \\
        &SemFew-Trans & Ours & Swin-T & \textbf{78.94 ± 0.66} & \textbf{86.49 ± 0.50} & \textbf{82.37 ± 0.77} & \textbf{89.89 ± 0.52} \\
        \bottomrule
    \end{tabular}
    \caption{Results (\%) on MiniImageNet and TieredImageNet. The ± shows 95\% confidence intervals. The best results are shown in \textbf{bold}.}
    \label{tab:table1}
    \vspace{-3mm}
\end{table*}
\section{Experiments}
We exhibit experimental results in this section, including benchmarks introduction, implementation details, and few-shot classification performance.

\subsection{Datasets}
The study evaluates the proposed method on six established FSL datasets: MiniImageNet \cite{vinyals2016matching}, TieredImageNet \cite{ren2018meta}, CIFAR-FS \cite{lee2019meta}, FC100 \cite{oreshkin2018tadam}, Places \cite{zhou2017places}, and CUB \cite{wah2011caltech}. \textbf{MiniImageNet} and \textbf{TieredImageNet} are both subsets of ImageNet \cite{deng2009imagenet}. MiniImageNet consists of 64 training classes, 16 validating classes, and 20 testing classes. TieredImageNet encompasses 351 training classes, 97 validating classes, and 160 testing classes. \textbf{CIFAR-FS} and \textbf{FC100} are derived from the CIFAR-100 \cite{krizhevsky2009learning}. CIFAR-FS employs a random partitioning strategy with 64 training classes, 16 validating classes, and 20 testing classes. FC100 introduces a unique superclass partitioning approach, where the training set comprises 12 superclasses, namely 60 classes, while both the validation and test sets include four superclasses, totaling 20 classes. \textbf{Places} and \textbf{Caltech-UCSD Birds-200-2011 (CUB)} are datasets for cross-domain scenario testing. In a cross-domain evaluation \cite{tseng2020cross}, the model is initially trained on MiniImageNet's base split and tested on the corresponding test split.

\begin{table*}[ht!]
    \centering
    \begin{tabular}{l c c c c c c}
    \toprule
        \multirow{2}{*}{\textbf{Method}}
         & \multirow{2}{*}{\textbf{Venue}} & \multirow{2}{*}{\textbf{Backbone}} & \multicolumn{2}{c}{\textbf{CIFAR-FS}} & \multicolumn{2}{c}{\textbf{FC100}}\\ 
        \multirow{2}{*}{} & \multirow{2}{*}{} & \multirow{2}{*}{} & \textbf{5-way 1-shot} & \textbf{5-way 5-shot}  & \textbf{5-way 1-shot} & \textbf{5-way 5-shot}\\ 
        \midrule
        ProtoNet \cite{snell2017prototypical} & NeurIPS'17 & ResNet-12 & 72.20 ± 0.70 & 83.50 ± 0.50 & 41.54 ± 0.76 & 57.08 ± 0.76\\ 
        TADAM \cite{oreshkin2018tadam} & NeurIPS'18 & ResNet-12 & - & - & 40.10 ± 0.40 & 56.10 ± 0.40 \\
        MetaOptNet \cite{lee2019meta} & CVPR'19 & ResNet-12 & 72.80 ± 0.70 & 84.30 ± 0.50 & 47.20 ± 0.60 & 55.50 ± 0.60\\
        MABAS \cite{kim2020model} & ECCV'20 & ResNet-12 & 73.51 ± 0.92 & 85.65 ± 0.65 & 42.31 ± 0.75 & 58.16 ± 0.78 \\
        RFS \cite{tian2020rethinking} & ECCV'20 & ResNet-12 & 71.50 ± 0.80 & 86.00 ± 0.50 & 42.60 ± 0.70 & 59.10 ± 0.60 \\
        SUN \cite{dong2022self} & ECCV'22 & ViT-S & 78.37 ± 0.46 & 88.84 ± 0.32 & - & -\\
        FewTURE \cite{hiller2022rethinking} & NeurIPS'22 & Swin-T & 77.76 ± 0.81 & 88.90 ± 0.59 & 47.68 ± 0.78 & 63.81 ± 0.75 \\
        Meta-AdaM \cite{metaAdaM} & NeurIPS'23 & ResNet-12 & - & - & 41.12 ± 0.49 & 56.14 ± 0.49 \\
        \midrule
        SP-CLIP \cite{chen2023semantic} & CVPR'23 & Visformer-T & 82.18 ± 0.40 & 88.24 ± 0.32 & 48.53 ± 0.38 & 61.55 ± 0.41 \\
        \midrule
        SemFew & Ours & ResNet-12 & 83.65 ± 0.70 & 87.66 ± 0.60 & \textbf{54.36 ± 0.71} & 62.79 ± 0.74 \\  
        SemFew-Trans & Ours & Swin-T & \textbf{84.34 ± 0.67} & \textbf{89.11 ± 0.54} & 54.27 ± 0.77 & \textbf{65.02 ± 0.72} \\
        \bottomrule
    \end{tabular}
    \caption{Results (\%) on CIFAR-FS and FC100. The ± shows 95\% confidence intervals. The best results are shown in \textbf{bold}.}
    \label{tab:table2}
    \vspace{-3mm}
\end{table*}

\begin{table}[t]
    \centering
    \resizebox{\linewidth}{!}{
    \begin{tabular}{l c c c c c}
    \toprule
         \multirow{2}{*}{\textbf{Method}} &  \multirow{2}{*}{\textbf{Venue}} & \multicolumn{2}{c}{\textbf{CUB}} &  \multicolumn{2}{c}{\textbf{Places}}\\
         &  & \textbf{1-shot} & \textbf{5-shot} & \textbf{1-shot} & \textbf{5-shot}\\
        \midrule
        GNN \cite{satorras2018few} & \small{ICLR'18} & 45.69 & 62.25 & 53.10 & 70.84 \\
        S2M2 \cite{mangla2020charting} & \small{WACV'20} & 48.24 & 70.44 & - & - \\
        FT \cite{tseng2020cross} & \small{ICLR'20} & 47.47 & 66.98 & 55.77 & 73.94 \\
        ATA \cite{ijcai2021-149} & \small{IJCAI'21} & 45.00 & 66.22 & 53.57 & 75.48 \\
        AFA \cite{hu2022adversarial} & \small{ECCV'22} & 46.86 & 68.25 & 54.04 & 76.21 \\
        StyleAdv \cite{fu2023styleadv} & \small{CVPR'23} & 48.49 & 68.72 & 58.58 & \textbf{77.73} \\
        LDP-net \cite{zhou2023revisiting} & \small{CVPR'23} & 49.82 & 70.39 & 53.82 & 72.90 \\
        \midrule
        SemFew-Name & \small{Ours} & 57.58 & 72.26 & 63.22 & 74.54 \\
        SemFew & \small{Ours} & \textbf{59.07} & \textbf{72.47} & \textbf{64.01} & 74.70 \\
        \bottomrule
    \end{tabular}}
    \caption{Average results (\%) on cross-domain scenarios. SemFew-Name denotes that semantics are class names.}
    \label{tab:table5}
    \vspace{-5mm}
\end{table}

\begin{table*}[t]
    \centering
    \begin{tabular}{c | c c c | c c c}
    \toprule
        \multirow{2}{*}{\textbf{Method}} & \multicolumn{3}{c|}{\textbf{5-way 1-shot}} & \multicolumn{3}{c}{\textbf{5-way 5-shot}} \\
        & \textbf{MiniImageNet} & \textbf{CIFAR-FS} & \textbf{FC100} & \textbf{MiniImageNet} & \textbf{CIFAR-FS} & \textbf{FC100} \\
        \midrule
        $V \Rightarrow C$ & 66.27 ± 0.81 & 74.43 ± 0.90 & 44.72 ± 0.75 & 81.70 ± 0.55 & 86.74 ± 0.61 & 60.58 ± 0.76 \\
        $S \Rightarrow C$ & 76.77 ± 0.66 & 82.27 ± 0.71 & 52.12 ± 0.73 & 82.92 ± 0.48 & 87.42 ± 0.60 & 62.22 ± 0.75 \\ 
        \midrule
        $V+S \Rightarrow C$ & \textbf{77.63 ± 0.63} & \textbf{83.65 ± 0.70} & \textbf{54.36 ± 0.71} & \textbf{83.04 ± 0.48} & \textbf{87.66 ± 0.60} & \textbf{62.79 ± 0.74} \\   
        \bottomrule
    \end{tabular}
    \caption{Results (\%) on different alignment sources. $V$ means visual features, $S$ represents semantic features, $C$ stands for class prototypes.}
    \label{tab:table6}
\end{table*}

\begin{table}[t]
    \centering
    \resizebox{\linewidth}{!}{
    \begin{tabular}{c | c c c c c c c c}
    \toprule
        \multirow{2}{*}{\textbf{Center}} & \multicolumn{2}{c}{\textbf{MiniImageNet}} & \multicolumn{2}{c}{\textbf{TieredImageNet}} & \multicolumn{2}{c}{\textbf{CIFAR-FS}} & \multicolumn{2}{c}{\textbf{FC100}} \\
        & \textbf{1-shot} & \textbf{5-shot} & \textbf{1-shot} & \textbf{5-shot} & \textbf{1-shot} & \textbf{5-shot} & \textbf{1-shot} & \textbf{5-shot}\\
        \midrule
        Cluster & 77.16 & 82.94 & 78.36 & 85.19 & 83.39 & 87.61 & 53.30 & 62.22 \\
        Mean & 77.63 & 83.04 & 78.96 & 85.88 & 83.65 & 87.66 & 54.36 & 62.79 \\
        \bottomrule
    \end{tabular}}
    \caption{Average results (\%) on different prototypes. \textbf{Mean} represents the mean vector of sample clusters. \textbf{Cluster} means the prototype is determined by Clustering Algorithm.}
    \label{tab:table7}
\end{table}

\begin{table}[t]
    \centering  
    \resizebox{\linewidth}{!}{
    \begin{tabular}{c | c c c c c c c c}
    \toprule
        \multirow{2}{*}{\textbf{classifier}} & \multicolumn{2}{c}{\textbf{MiniImageNet}} & \multicolumn{2}{c}{\textbf{TieredImageNet}} & \multicolumn{2}{c}{\textbf{CIFAR-FS}} & \multicolumn{2}{c}{\textbf{FC100}} \\
        & \textbf{1-shot} & \textbf{5-shot} & \textbf{1-shot} & \textbf{5-shot} & \textbf{1-shot} & \textbf{5-shot} & \textbf{1-shot} & \textbf{5-shot}\\
        \midrule
        LR & 77.55 & 82.70 & 78.66 & 84.72 & 83.08 & 86.70 & 51.53 & 57.49 \\
        EU & 77.56 & 83.15 & 78.65 & 85.26 & 82.83 & 86.87 & 54.14 & 62.54 \\
        CO & 77.63 & 83.04 & 78.96 & 85.88 & 83.65 & 87.66 & 54.36 & 62.79 \\
        \bottomrule
    \end{tabular}}
    \caption{Average results (\%) on different classifiers. \textbf{LR}: linear logistic regression classifier. \textbf{EU}: euclidean-distance nearest prototype classifier. \textbf{CO}: cosine-distance nearest prototype classifier. }
    \label{tab:table8}
    \vspace{-1mm}
\end{table}

\subsection{Implementation Details}
\textbf{Architecture.}
We employ a pre-trained ResNet-12 \cite{he2016deep} and Swin-T \cite{liu2021swin} as the vision encoders. Additionally, we apply two types of text encoder: the text encoder from ViT-B/16 CLIP \cite{radford2021learning} where output dimension is 512, and the BERT-Base text encoder \cite{devlin2018bert} which has 768 output dimension. The network $h$ that combines visual and textual features consists of two fully connected layers and a LeakyReLU \cite{maas2013rectifier} activation function. The hidden layer has a dimension of 4,096. In the context of Semantic Evolution, the utilized Large Language Model is GPT-3.5-turbo \cite{ChatGPT}.

\textbf{Training Details.}
A two-step training procedure is performed in experiments. In step one, two kinds of vision encoder are pre-trained, the ResNet-12 \cite{he2016deep} follows the same pretraining settings as the FEAT \cite{ye2020few}, while the Swin-T \cite{liu2021swin} adopts the identical pretraining strategy as FewTURE \cite{hiller2022rethinking}. Next, the network $h$ is trained 50 epochs. This training is conducted with a batch size of 128 and employs an Adam optimizer \cite{kingma2014adam} with a learning rate of 1e-4.

\textbf{Evaluation protocol.}
The proposed method is evaluated under 5-way 1-shot and 5-way 5-shot settings. For each setting, 600 classification tasks are uniformly sampled from the novel set. In each task, there are 15 query samples per class. The mean and 95\% confidence interval of the accuracy are reported.

\subsection{Experimental Results}
\subsubsection{Few-Shot Classification}
The comparative results of our proposed approach against previous methods across four benchmarks are shown in \cref{tab:table1,tab:table2}. Specifically, in the 1-shot task, SemFew exhibits notable superiority, surpassing SP-CLIP \cite{chen2023semantic} \textbf{0.93\%-5.83\%} on four benchmarks. Recent studies \cite{chen2023semantic,hiller2022rethinking,dong2022self} have explored different Transformer architectures as backbones. Consequently, we conduct experiments with a Transformer backbone, resulting in SemFew-Trans. This variant further elevates performance, outperforming SP-CLIP by \textbf{2.16\%-6.63\%} in the 1-shot setting. In summary, SemFew-Trans achieves state-of-the-art results compared to both previous visual-based and semantic-based approaches across all 1-shot and 5-shot tasks. 

\subsubsection{Cross-Domain Classification}
To evaluate whether semantics serve as an effective bridge across novel tasks, we test SemFew on completely novel datasets. Such a task is also known as Cross-Domain Few-Shot Learning (CDFSL) \cite{tseng2020cross}, which involves extending the capacity to recognize novel categories to domains distinct from the training domain. All the results are reported in \cref{tab:table5}, where the experimental setup is the same as FT \cite{tseng2020cross}. 
\cref{tab:table5} indicates that SemFew significantly surpassing the state-of-the-art methods, where it outperforms best compared methods by \textbf{9.25}\% (1-shot) and \textbf{1.87}\% (5-shot) on CUB. It also surpasses the lastest method \cite{zhou2023revisiting} \textbf{10.19}\% (1-shot) and \textbf{1.80}\% (5-shot) on Places.
Conversely, SemFew-Name, which adopts a template (\ie, \textit{The Photo of a bird called \{Class Name\}}) as the semantic, performs worse than SemFew. This observation highlights the inadequacy of using solely names as semantics.

\subsection{Ablation Studies}
We conduct ablation experiments on three datasets, \ie, MiniImageNet, CIFAR-FS, and FC100, and all experiments are performed using the ResNet-12 network architecture.

\subsubsection{Alignment Source}\label{learningalgo}
Given that our framework is primarily responsible for translating visual samples $V$ and semantics descriptions $S$ into ground-truth class prototype (\ie, $V + S \Rightarrow C$), we conduct ablation studies by removing $S$ or $V$. The results can be found in \cref{tab:table6}. First, we remove semantics and force pure visual samples to reconstruct prototypes (\ie, $V \Rightarrow C$). The result shows that it gets the worst performance, indicating that reconstructing a prototype from pure visual features is challenging. Then, we abandon visual features to implement our proposed principle in \cref{sec3.3:semfew} (\ie, $S \Rightarrow C$). As we can see, it achieves substantial performance gains, reflecting the rich class representative features within semantics. However, our method combining visual and semantic acquires the best performance, which indicates the complementarity between visual and semantic modalities does enhance the robustness of the reconstructed prototype.

\subsubsection{Prototype and Classifier Selection Strategy}
\cref{tab:table7} displays the performance associated with various learning targets. Results reveal that employing the mean vector as a prototype yields slightly better results compared to prototypes identified by the clustering algorithm \cite{hartigan1979algorithm}. This suggests that the mean vector consolidates representative visual information, making it a highly effective class prototype. As for the classification head design, we present the results in \cref{tab:table8}, which indicates the use of cosine distance as the metric to achieve nearest neighbor classification gets the best results, euclidean distance gets similar but lower outcomes, and linear logistic classifier gets the worst performance.

\subsubsection{Semantic Evolution}
\begin{figure}[t]
    \begin{center}
    \begin{subfigure}[b]{.49\linewidth}
        \centering
        \includegraphics[width=\linewidth]{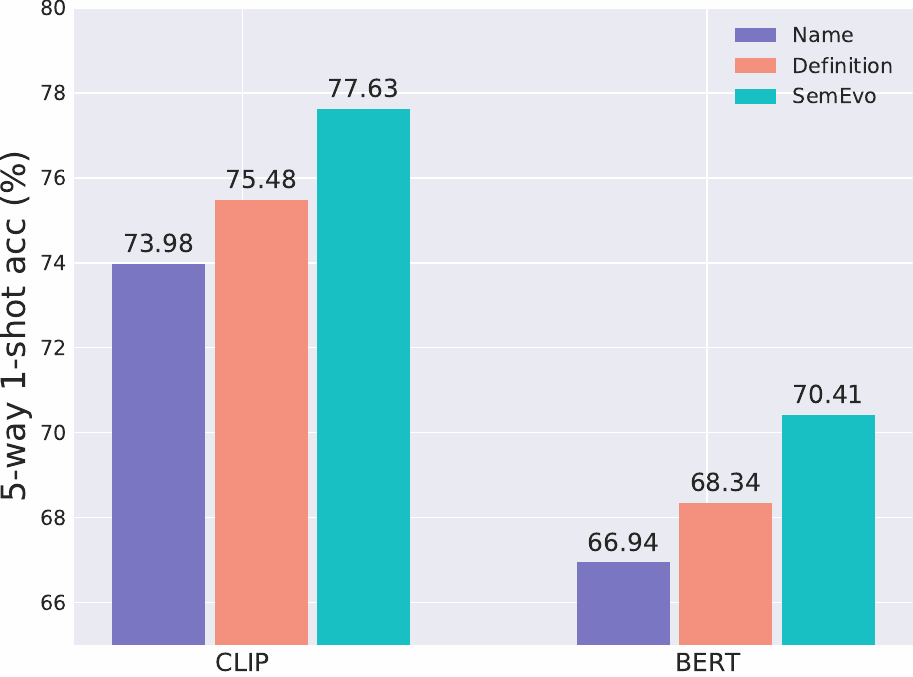} 
        \caption{MiniImageNet, 1-shot}
        \label{sfig4:a}
    \end{subfigure}
    \begin{subfigure}[b]{.49\linewidth}
        \centering
        \includegraphics[width=\linewidth]{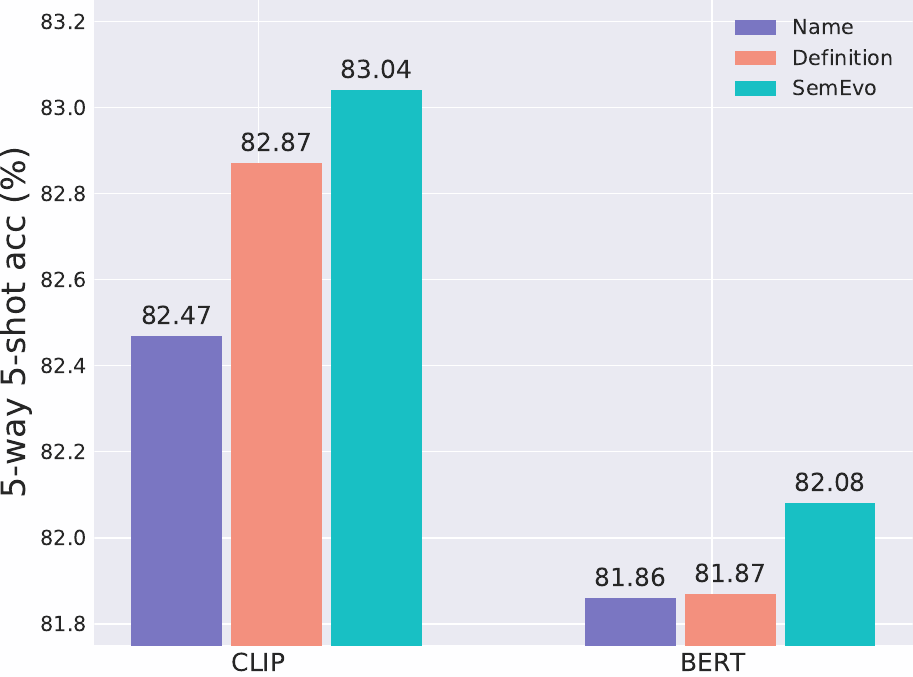} 
        \caption{MiniImageNet, 5-shot}
        \label{sfig4:b}
    \end{subfigure} \\
    \begin{subfigure}[b]{.49\linewidth}
        \centering
        \includegraphics[width=\linewidth]{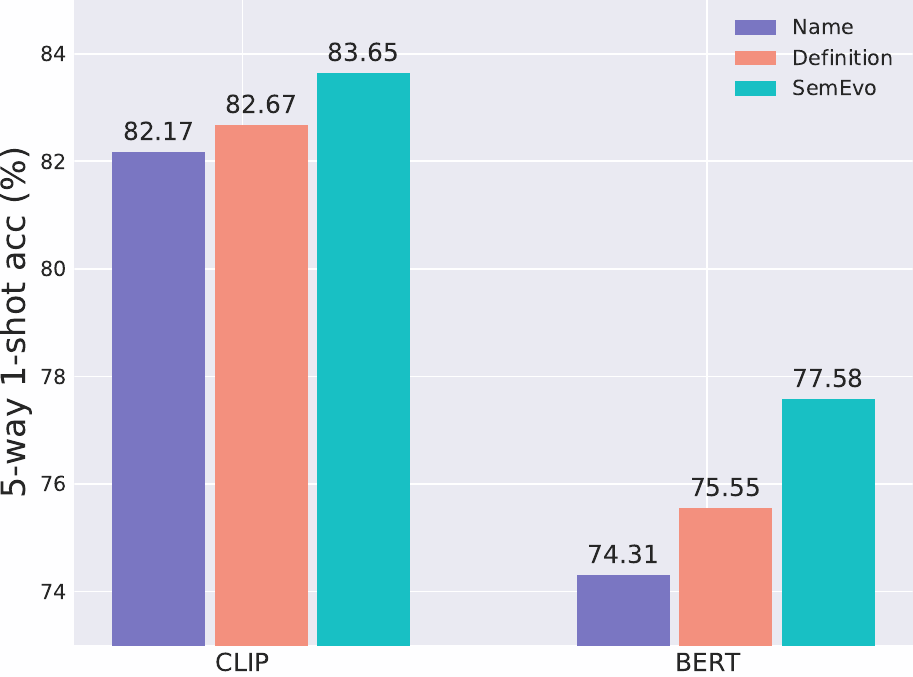}
        \caption{CIFAR-FS, 1-shot}
        \label{sfig4:c}
    \end{subfigure} 
    \begin{subfigure}[b]{.49\linewidth}
        \centering
        \includegraphics[width=\linewidth]{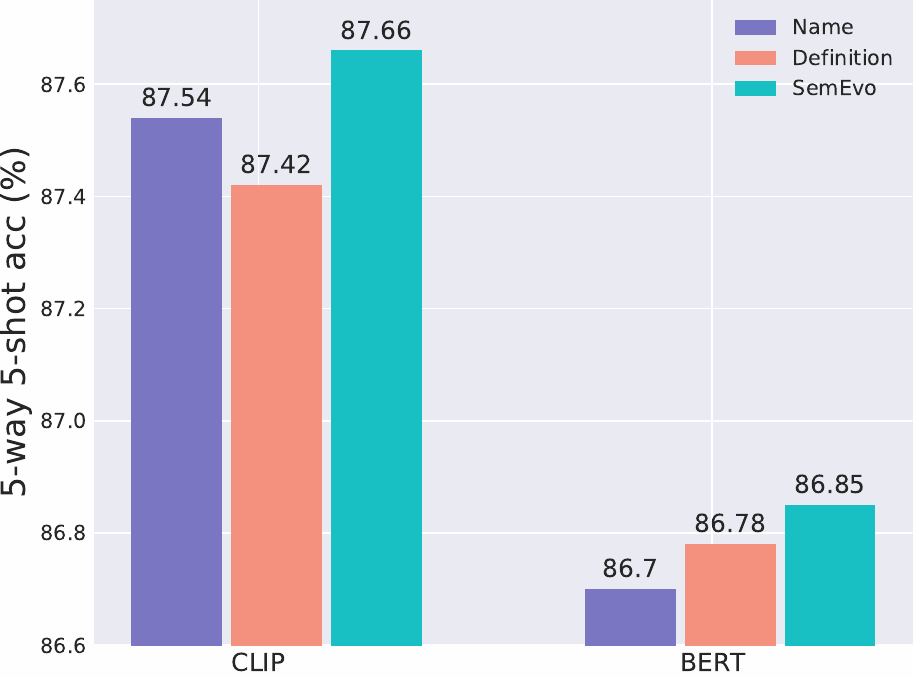}
        \caption{CIFAR-FS, 5-shot}
        \label{sfig4:d}
    \end{subfigure}\\
        \begin{subfigure}[b]{.49\linewidth}
        \centering
        \includegraphics[width=\linewidth]{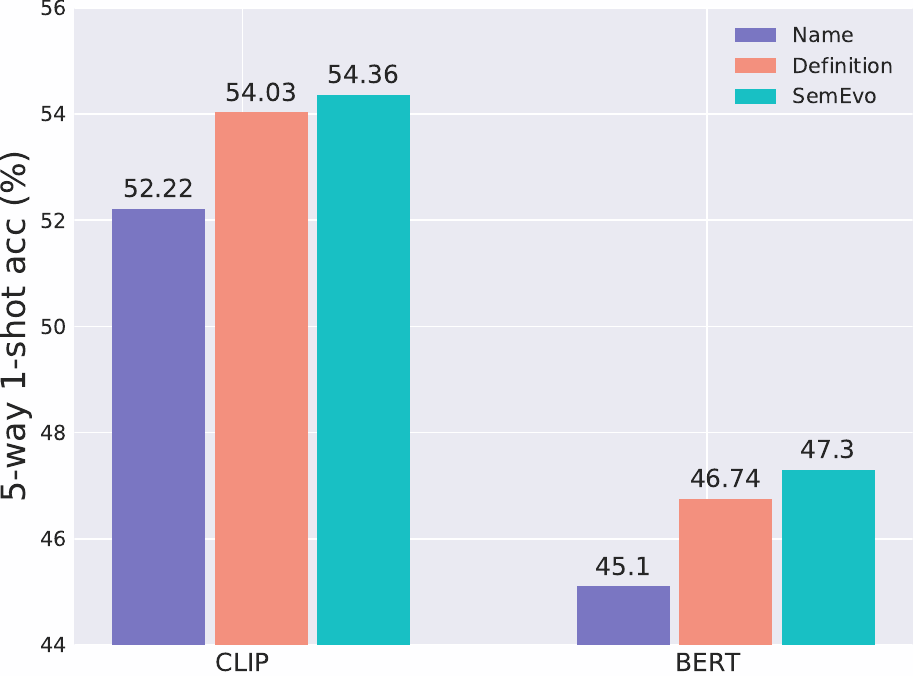}
        \caption{FC100, 1-shot}
        \label{sfig4:e}
    \end{subfigure}
        \begin{subfigure}[b]{.49\linewidth}
        \centering
        \includegraphics[width=\linewidth]{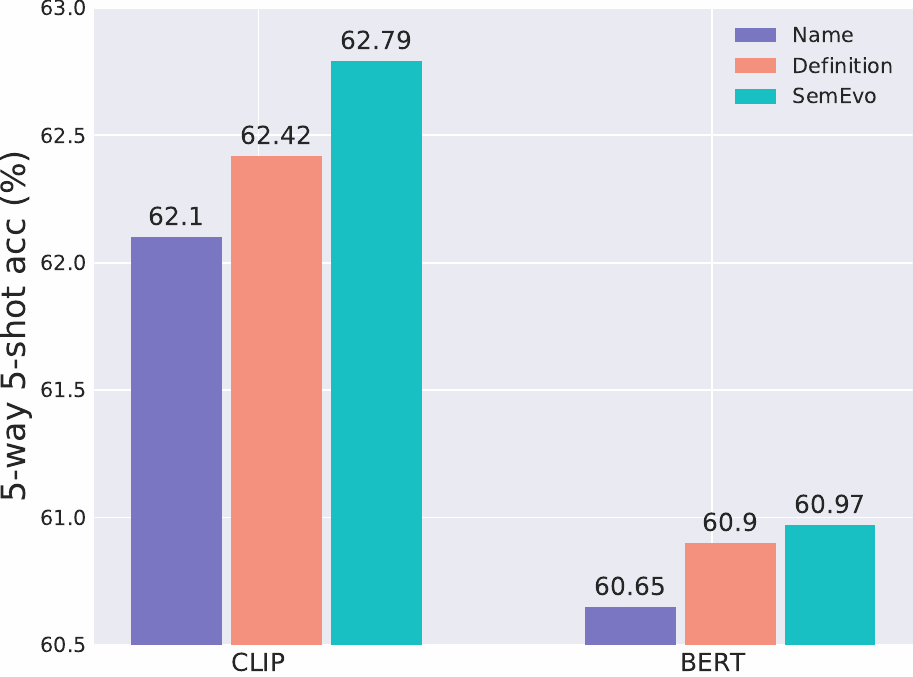}
        \caption{FC100, 5-shot}
        \label{sfig4:f}
    \end{subfigure}
    \caption{Average results (\%) on different semantics.}
    \label{fig:fig4}
    \vspace{-7mm}
    \end{center}
\end{figure}

We use different kinds of semantics and text encoders to evaluate the influence of semantics, where results can be found in \cref{fig:fig4}. In most scenarios, our model attains optimal results when aided by paraphrased semantics, and the model trained on simple definitions secures the second-best performance. This trend is reasonable since paraphrased descriptions contain more knowledge than simple definitions and class names. Between the comparison of different text encoders, the model trained on CLIP encoded semantics greatly outperforms the model trained on BERT encoded semantics. This is because CLIP is trained in a multi-modal manner, rendering it better capable of understanding images. Although a multi-modal text encoder has been widely adopted in FSL research, our model obtains competitive results on BERT, demonstrating that uni-modal semantics can also achieve excellent results.

\begin{figure*}[t]
    \begin{center}
    \begin{subfigure}[b]{.23\linewidth}
        \centering
        \begin{mdframed}
        \includegraphics[width=\linewidth]{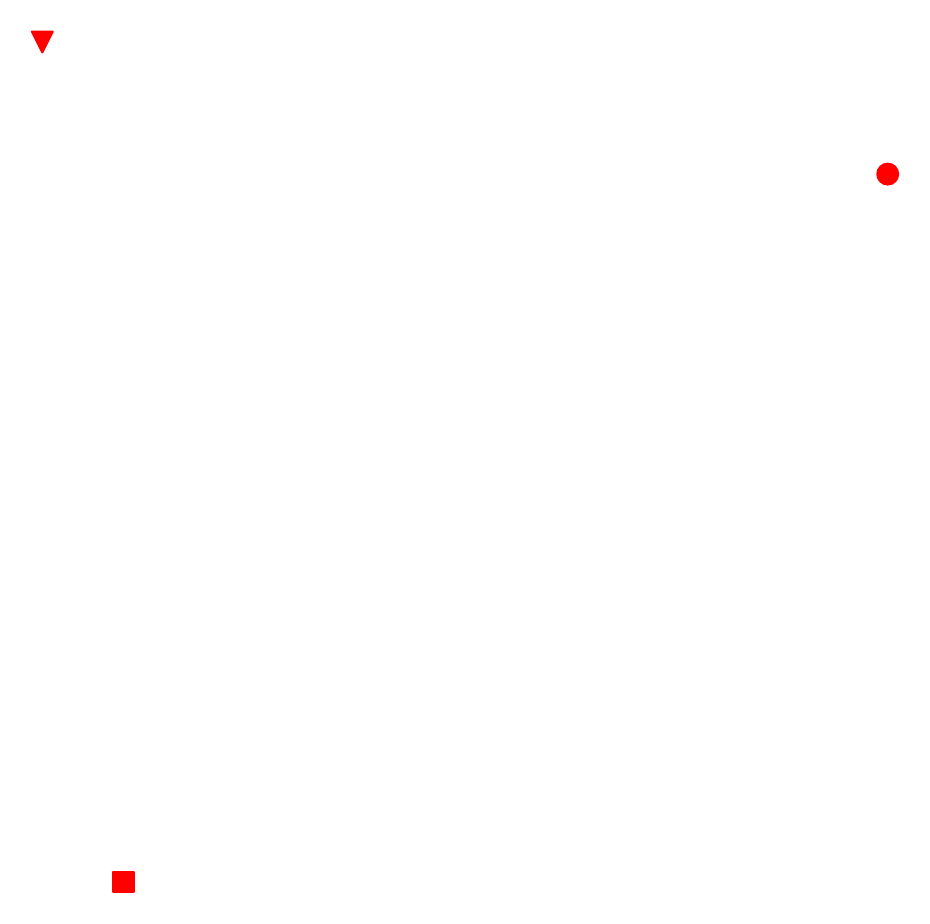} 
        \end{mdframed}
        \caption{Support}
        \label{sfig5:a}
    \end{subfigure}\hfill
    \begin{subfigure}[b]{.23\linewidth}
        \centering
        \begin{mdframed}
        \includegraphics[width=\linewidth]{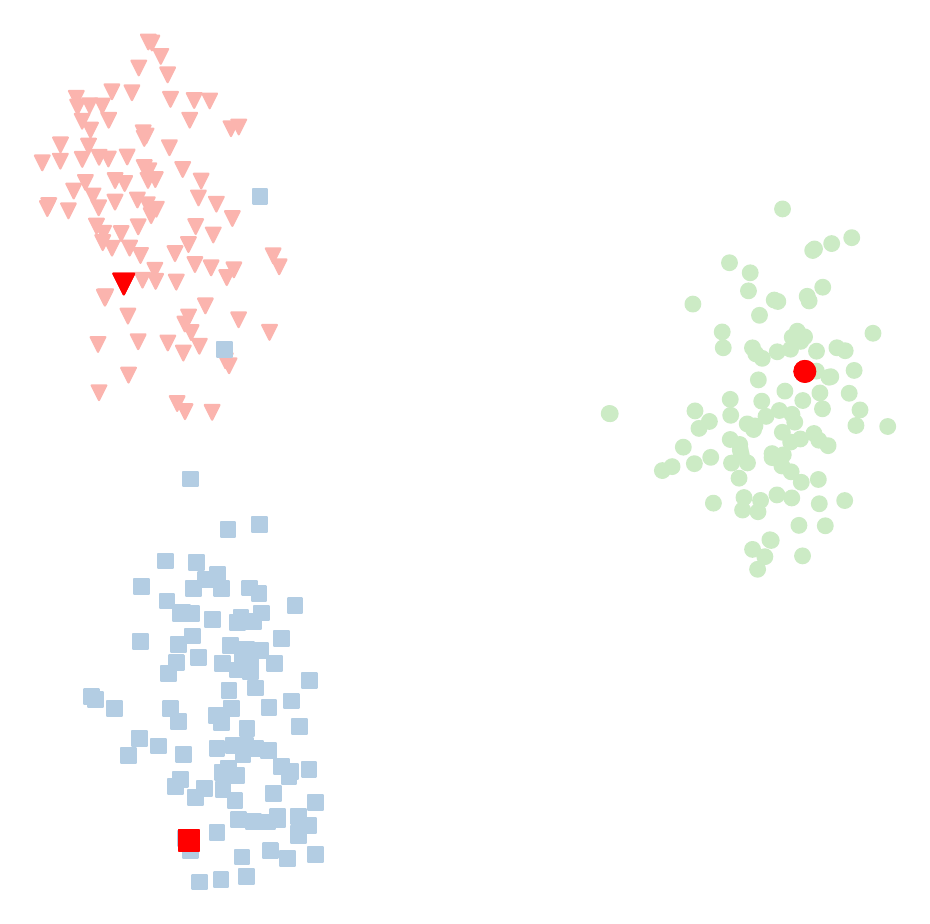} 
        \end{mdframed}
        \caption{Query}
        \label{sfig5:b}
    \end{subfigure}\hfill
    \begin{subfigure}[b]{.23\linewidth}
        \centering
        \begin{mdframed}
        \includegraphics[width=\linewidth]{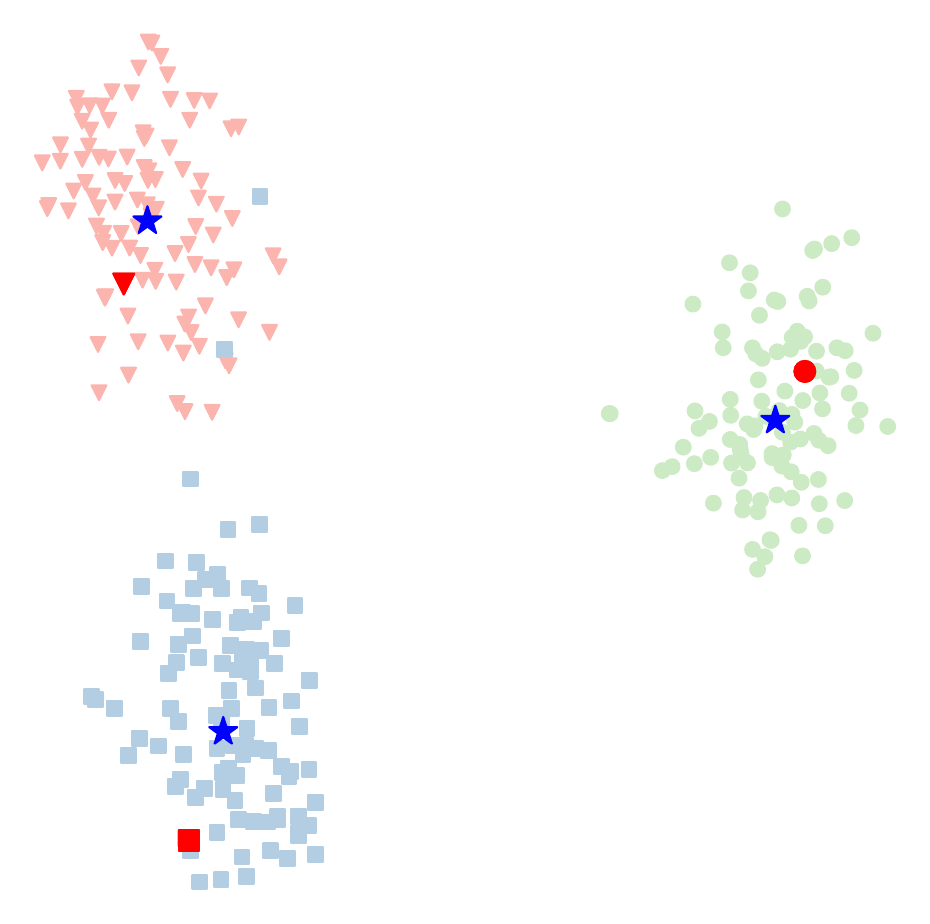}
        \end{mdframed}
        \caption{Real Prototype}
        \label{sfig5:c}
    \end{subfigure}\hfill
    \begin{subfigure}[b]{.23\linewidth}
        \centering
        \begin{mdframed}
        \includegraphics[width=\linewidth]{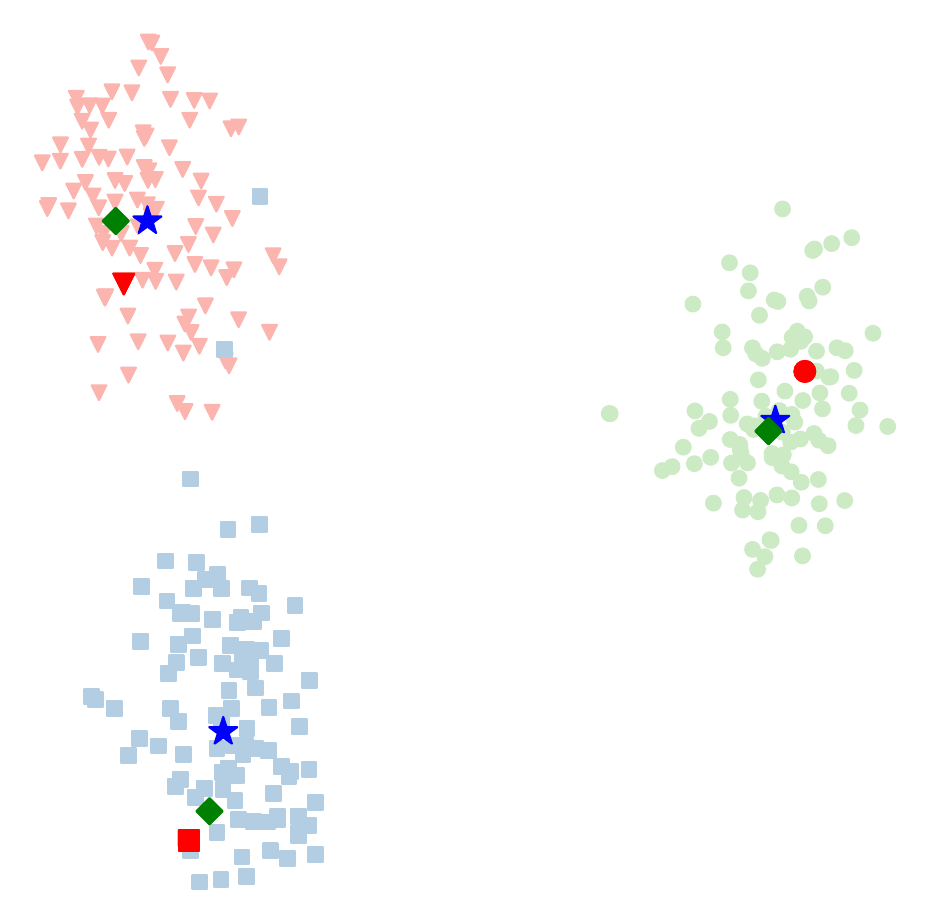} 
        \end{mdframed}
        \caption{Reconstructed Prototype}
        \label{sfig5:d}
    \end{subfigure}
    \caption{Visualization results on the MiniImageNet dataset. Different colors or shapes represent different classes. The $\star$ represents the class prototypes, and the $\diamondsuit$ denotes the prototypes reconstructed by our method.}
    \label{fig:fig5}
    \end{center}
    \vspace{-5mm}
\end{figure*}

\begin{figure}[t]
    \begin{center}
    \begin{subfigure}[b]{.49\linewidth}
        \centering
        \includegraphics[width=\linewidth]{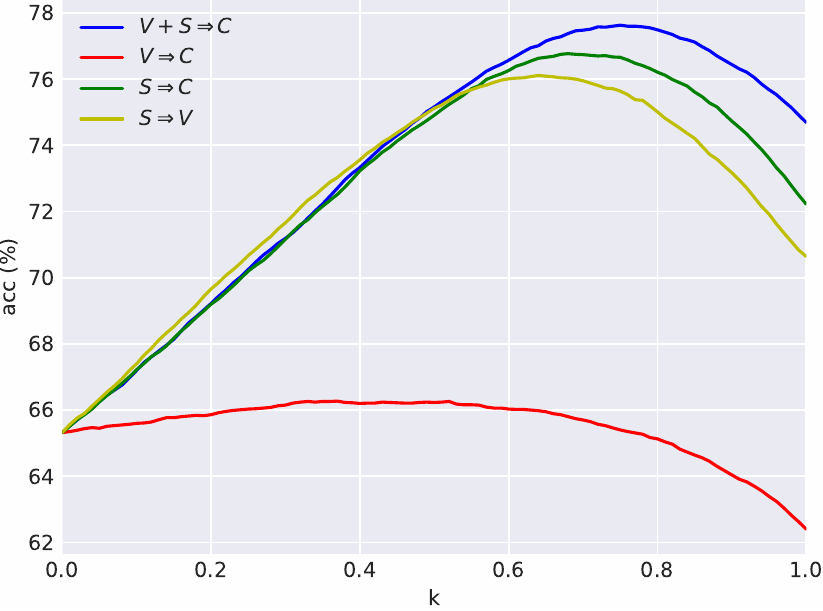} 
        \caption{MiniImageNet, CLIP}
        \label{sfig6:a}
    \end{subfigure}\
    \begin{subfigure}[b]{.49\linewidth}
        \centering
        \includegraphics[width=\linewidth]{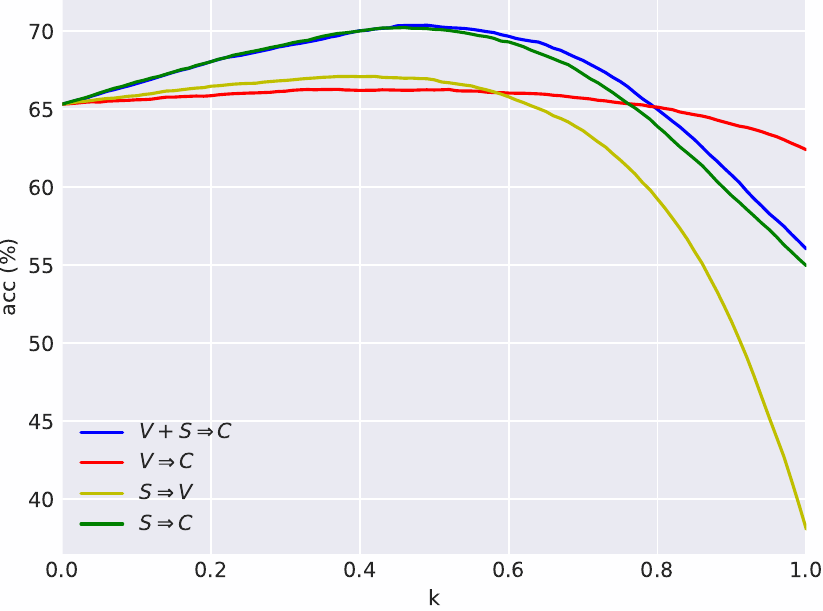} 
        \caption{MiniImageNet, BERT}
        \label{sfig6:b}
    \end{subfigure}\\
    \begin{subfigure}[b]{.49\linewidth}
        \centering
        \includegraphics[width=\linewidth]{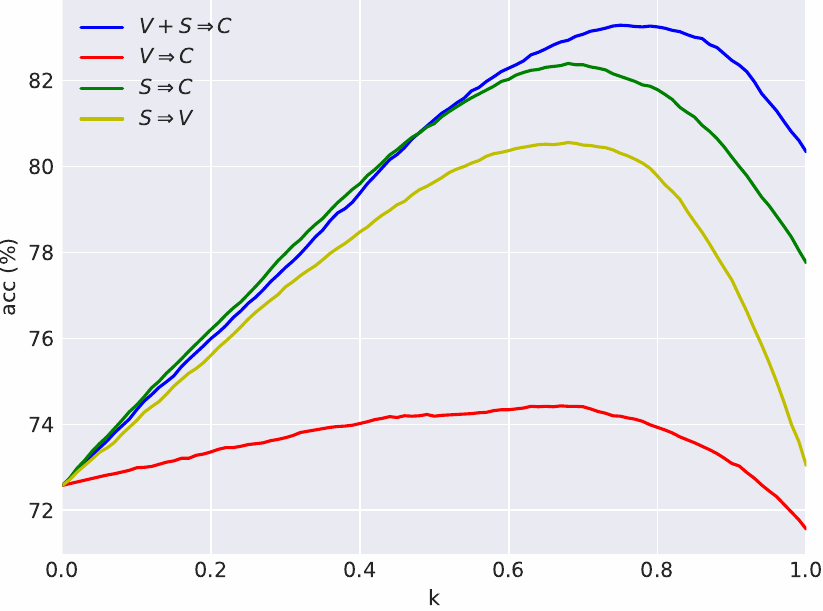} 
        \caption{CIFAR-FS, CLIP}
        \label{sfig6:c}
    \end{subfigure} 
    \begin{subfigure}[b]{.49\linewidth}
        \centering
        \includegraphics[width=\linewidth]{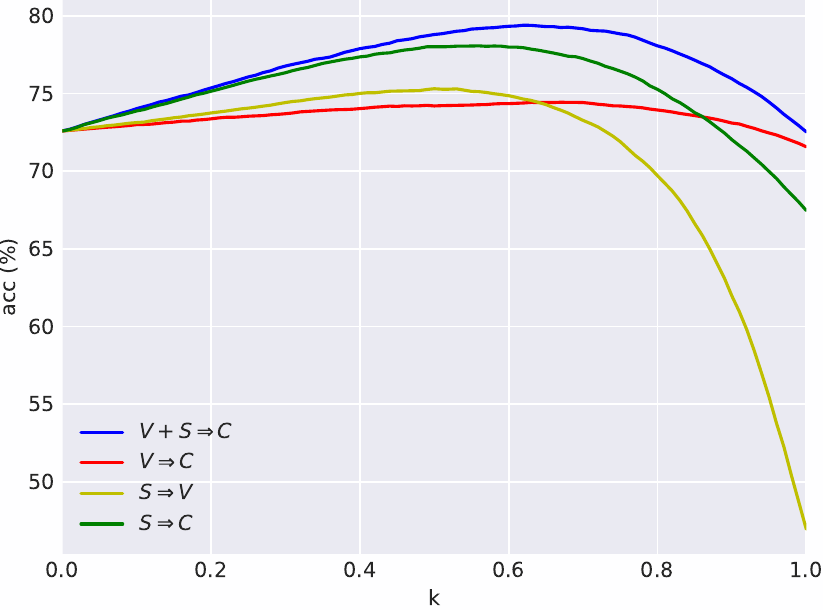} 
        \caption{CIFAR-FS, BERT}
        \label{sfig6:d}
    \end{subfigure} \\
    \begin{subfigure}[b]{.49\linewidth}
        \centering
        \includegraphics[width=\linewidth]{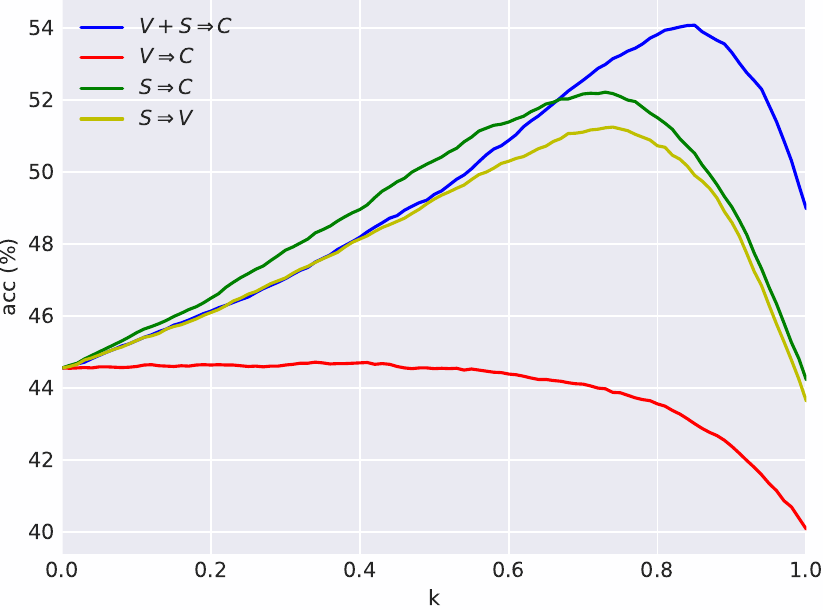}
        \caption{FC100, CLIP}
        \label{sfig6:e}
    \end{subfigure} 
    \begin{subfigure}[b]{.49\linewidth}
        \centering
        \includegraphics[width=\linewidth]{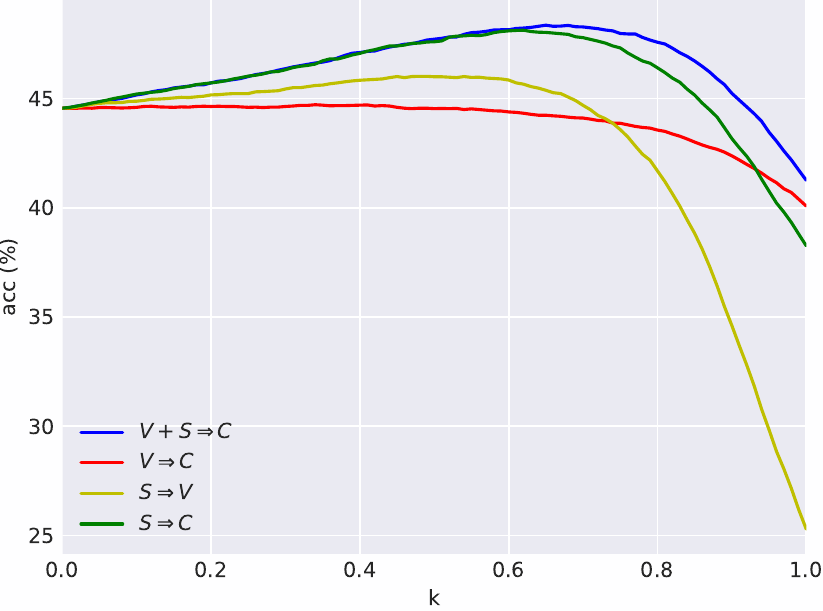}
        \caption{FC100, BERT}
        \label{sfig6:f}
    \end{subfigure} 
    \caption{Average results (\%) on different fusion factor $k$.}
    \label{fig:fig6}
    \end{center}
    \vspace{-7mm}
\end{figure}

\subsection{Visualization Analysis}
We visualize visual space features through the t-SNE algorithm \cite{van2008visualizing}, where the results are presented in \cref{fig:fig5}. We select three novel classes on the MiniImageNet dataset, and then randomly sampled three support samples and one hundred query samples, where the distribution of support and query samples in visual space can be found in \cref{sfig5:a,sfig5:b}. In \cref{sfig5:c}, we put support samples, \ie, red dots, mean vector of each class, \ie, blue stars, and query samples in the same picture. As we can see, support samples are close to the cluster edge rather than the mean center, which leads to difficulties in correctly classifying most samples in the query set. However, after being transformed by SemFew, the reconstructed prototypes are closer to the mean center of each cluster compared to the support set samples. This phenomenon demonstrates that our method can reasonably transform visual samples into class prototypes corresponding to their respective categories.

\subsection{Fusion Factor Analysis}
The fusion factor, denoted as $k$ in \cref{con:convex}, significantly influences performance. To demonstrate, we systematically vary the value of $k$ between 0 and 1, using a step size of 0.01, and present the accuracy trends across three benchmarks in \cref{fig:fig6}. Notably, our model, \ie, blue line, gets the highest accuracy when $k$ is large. Given that $k$ represents the percentage of the reconstructed prototype $r_t$ during fusion in \cref{con:convex}, the experiment highlights that our model produces high-quality prototypes. This observation is further supported by the accuracy comparison when $k = 1$. In such cases, our model consistently outperforms others, providing strong evidence for the effectiveness of our reconstructed prototypes. 

Additionally, we add $S \Rightarrow V$ experiment for comparison, which generates visual samples rather than prototypes through a Conditional Generative Adversarial Network \cite{mirza2014conditional}. This experiment allows us to explore whether we can build robust prototypes indirectly by expanding the sample of the support set. As we can see, $S \Rightarrow V$, \ie, yellow line, gets relatively low performance in all settings. The experiments indicate that producing the prototype by expanding then averaging the support set is not as effective as directly generating class prototypes.

\section{Conclusion} 
To unleash the potential of semantics in FSL, we propose the Semantic Evolution to automatically generate high-quality semantics from simple class names. Then, we design a simple Semantic Alignment Network to transform the concatenated visual features and high-quality semantics features into robust class prototypes for classification. The experimental results show that with high-quality semantics, the basic network can easily achieve greater performance compared to other state-of-the-art methods. 

\noindent \textbf{Acknowledgement.} This work was supported by 2023YFF1204901, NSFC-62076172, 2021DQ02-0903, 2023YFG0116.

{
    \small
    \bibliographystyle{ieeenat_fullname}
    \bibliography{main}

\begin{thebibliography}{65}
\providecommand{\natexlab}[1]{#1}
\providecommand{\url}[1]{\texttt{#1}}
\expandafter\ifx\csname urlstyle\endcsname\relax
  \providecommand{\doi}[1]{doi: #1}\else
  \providecommand{\doi}{doi: \begingroup \urlstyle{rm}\Url}\fi

\bibitem[Andrychowicz et~al.(2016)Andrychowicz, Denil, Gomez, Hoffman, Pfau, Schaul, Shillingford, and De~Freitas]{andrychowicz2016learning}
Marcin Andrychowicz, Misha Denil, Sergio Gomez, Matthew~W Hoffman, David Pfau, Tom Schaul, Brendan Shillingford, and Nando De~Freitas.
\newblock Learning to learn by gradient descent by gradient descent.
\newblock In \emph{NeurIPS}, 2016.

\bibitem[Bateni et~al.(2020)Bateni, Goyal, Masrani, Wood, and Sigal]{bateni2020improved}
Peyman Bateni, Raghav Goyal, Vaden Masrani, Frank Wood, and Leonid Sigal.
\newblock Improved few-shot visual classification.
\newblock In \emph{CVPR}, pages 14493--14502, 2020.

\bibitem[Biederman(1987)]{biederman1987recognition}
Irving Biederman.
\newblock Recognition-by-components: a theory of human image understanding.
\newblock \emph{Psychol Rev}, 94\penalty0 (2):\penalty0 115, 1987.

\bibitem[Brown et~al.(2020)Brown, Mann, Ryder, Subbiah, Kaplan, Dhariwal, Neelakantan, Shyam, Sastry, Askell, et~al.]{brown2020language}
Tom Brown, Benjamin Mann, Nick Ryder, Melanie Subbiah, Jared~D Kaplan, Prafulla Dhariwal, Arvind Neelakantan, Pranav Shyam, Girish Sastry, Amanda Askell, et~al.
\newblock Language models are few-shot learners.
\newblock \emph{NeurIPS}, 33:\penalty0 1877--1901, 2020.

\bibitem[Chen et~al.(2023)Chen, Si, Zhang, Wang, Wang, and Tan]{chen2023semantic}
Wentao Chen, Chenyang Si, Zhang Zhang, Liang Wang, Zilei Wang, and Tieniu Tan.
\newblock Semantic prompt for few-shot image recognition.
\newblock In \emph{CVPR}, pages 23581--23591, 2023.

\bibitem[Chen et~al.(2021)Chen, Liu, Xu, Darrell, and Wang]{chen2021meta}
Yinbo Chen, Zhuang Liu, Huijuan Xu, Trevor Darrell, and Xiaolong Wang.
\newblock Meta-baseline: Exploring simple meta-learning for few-shot learning.
\newblock In \emph{ICCV}, pages 9062--9071, 2021.

\bibitem[Cheng et~al.(2023)Cheng, Yang, Zhou, Guo, and Wen]{cheng2023frequency}
Hao Cheng, Siyuan Yang, Joey~Tianyi Zhou, Lanqing Guo, and Bihan Wen.
\newblock Frequency guidance matters in few-shot learning.
\newblock In \emph{ICCV}, pages 11814--11824, 2023.

\bibitem[Deng et~al.(2009)Deng, Dong, Socher, Li, Li, and Fei-Fei]{deng2009imagenet}
Jia Deng, Wei Dong, Richard Socher, Li-Jia Li, Kai Li, and Li Fei-Fei.
\newblock Imagenet: A large-scale hierarchical image database.
\newblock In \emph{CVPR}, pages 248--255. Ieee, 2009.

\bibitem[Devlin et~al.(2018)Devlin, Chang, Lee, and Toutanova]{devlin2018bert}
Jacob Devlin, Ming-Wei Chang, Kenton Lee, and Kristina Toutanova.
\newblock Bert: Pre-training of deep bidirectional transformers for language understanding.
\newblock \emph{arXiv preprint arXiv:1810.04805}, 2018.

\bibitem[Doersch et~al.(2020)Doersch, Gupta, and Zisserman]{doersch2020crosstransformers}
Carl Doersch, Ankush Gupta, and Andrew Zisserman.
\newblock Crosstransformers: spatially-aware few-shot transfer.
\newblock In \emph{NeurIPS}, pages 21981--21993, 2020.

\bibitem[Dong et~al.(2022)Dong, Zhou, Yan, and Zuo]{dong2022self}
Bowen Dong, Pan Zhou, Shuicheng Yan, and Wangmeng Zuo.
\newblock Self-promoted supervision for few-shot transformer.
\newblock In \emph{ECCV}, pages 329--347. Springer, 2022.

\bibitem[Elsken et~al.(2020)Elsken, Staffler, Metzen, and Hutter]{elsken2020meta}
Thomas Elsken, Benedikt Staffler, Jan~Hendrik Metzen, and Frank Hutter.
\newblock Meta-learning of neural architectures for few-shot learning.
\newblock In \emph{CVPR}, pages 12365--12375, 2020.

\bibitem[Finn et~al.(2017)Finn, Abbeel, and Levine]{finn2017model}
Chelsea Finn, Pieter Abbeel, and Sergey Levine.
\newblock Model-agnostic meta-learning for fast adaptation of deep networks.
\newblock In \emph{Int. Conf. Mach. Learn.}, pages 1126--1135. PMLR, 2017.

\bibitem[Fu et~al.(2023)Fu, Xie, Fu, and Jiang]{fu2023styleadv}
Yuqian Fu, Yu Xie, Yanwei Fu, and Yu-Gang Jiang.
\newblock Styleadv: Meta style adversarial training for cross-domain few-shot learning.
\newblock In \emph{CVPR}, pages 24575--24584, 2023.

\bibitem[Gidaris et~al.(2019)Gidaris, Bursuc, Komodakis, P{\'e}rez, and Cord]{gidaris2019boosting}
Spyros Gidaris, Andrei Bursuc, Nikos Komodakis, Patrick P{\'e}rez, and Matthieu Cord.
\newblock Boosting few-shot visual learning with self-supervision.
\newblock In \emph{ICCV}, pages 8059--8068, 2019.

\bibitem[Hao et~al.(2023)Hao, He, Liu, Wu, Tao, and Cheng]{hao2023class}
Fusheng Hao, Fengxiang He, Liu Liu, Fuxiang Wu, Dacheng Tao, and Jun Cheng.
\newblock Class-aware patch embedding adaptation for few-shot image classification.
\newblock In \emph{ICCV}, pages 18905--18915, 2023.

\bibitem[Hartigan and Wong(1979)]{hartigan1979algorithm}
John~A Hartigan and Manchek~A Wong.
\newblock Algorithm as 136: A k-means clustering algorithm.
\newblock \emph{J R Stat Soc Ser C Appl Stat}, 28\penalty0 (1):\penalty0 100--108, 1979.

\bibitem[He et~al.(2016)He, Zhang, Ren, and Sun]{he2016deep}
Kaiming He, Xiangyu Zhang, Shaoqing Ren, and Jian Sun.
\newblock Deep residual learning for image recognition.
\newblock In \emph{CVPR}, pages 770--778, 2016.

\bibitem[Hiller et~al.(2022)Hiller, Ma, Harandi, and Drummond]{hiller2022rethinking}
Markus Hiller, Rongkai Ma, Mehrtash Harandi, and Tom Drummond.
\newblock Rethinking generalization in few-shot classification.
\newblock \emph{NeurIPS}, 35:\penalty0 3582--3595, 2022.

\bibitem[Hou et~al.(2019)Hou, Chang, Ma, Shan, and Chen]{hou2019cross}
Ruibing Hou, Hong Chang, Bingpeng Ma, Shiguang Shan, and Xilin Chen.
\newblock Cross attention network for few-shot classification.
\newblock In \emph{NeurIPS}, 2019.

\bibitem[Hu and Ma(2022)]{hu2022adversarial}
Yanxu Hu and Andy~J Ma.
\newblock Adversarial feature augmentation for cross-domain few-shot classification.
\newblock In \emph{ECCV}, pages 20--37. Springer, 2022.

\bibitem[Jamal and Qi(2019)]{jamal2019task}
Muhammad~Abdullah Jamal and Guo-Jun Qi.
\newblock Task agnostic meta-learning for few-shot learning.
\newblock In \emph{CVPR}, pages 11719--11727, 2019.

\bibitem[Kim et~al.(2020)Kim, Kim, and Kim]{kim2020model}
Jaekyeom Kim, Hyoungseok Kim, and Gunhee Kim.
\newblock Model-agnostic boundary-adversarial sampling for test-time generalization in few-shot learning.
\newblock In \emph{ECCV}, pages 599--617. Springer, 2020.

\bibitem[Kingma and Ba(2014)]{kingma2014adam}
Diederik~P Kingma and Jimmy Ba.
\newblock Adam: A method for stochastic optimization.
\newblock \emph{arXiv preprint arXiv:1412.6980}, 2014.

\bibitem[Koch et~al.(2015)Koch, Zemel, Salakhutdinov, et~al.]{koch2015siamese}
Gregory Koch, Richard Zemel, Ruslan Salakhutdinov, et~al.
\newblock Siamese neural networks for one-shot image recognition.
\newblock In \emph{ICML Deep Learn. workshop}. Lille, 2015.

\bibitem[Krizhevsky et~al.(2009)Krizhevsky, Hinton, et~al.]{krizhevsky2009learning}
Alex Krizhevsky, Geoffrey Hinton, et~al.
\newblock Learning multiple layers of features from tiny images.
\newblock 2009.

\bibitem[Lee et~al.(2019)Lee, Maji, Ravichandran, and Soatto]{lee2019meta}
Kwonjoon Lee, Subhransu Maji, Avinash Ravichandran, and Stefano Soatto.
\newblock Meta-learning with differentiable convex optimization.
\newblock In \emph{CVPR}, pages 10657--10665, 2019.

\bibitem[Li et~al.(2020)Li, Huang, Lan, Feng, Li, and Wang]{li2020boosting}
Aoxue Li, Weiran Huang, Xu Lan, Jiashi Feng, Zhenguo Li, and Liwei Wang.
\newblock Boosting few-shot learning with adaptive margin loss.
\newblock In \emph{CVPR}, pages 12576--12584, 2020.

\bibitem[Li et~al.(2019)Li, Eigen, Dodge, Zeiler, and Wang]{li2019finding}
Hongyang Li, David Eigen, Samuel Dodge, Matthew Zeiler, and Xiaogang Wang.
\newblock Finding task-relevant features for few-shot learning by category traversal.
\newblock In \emph{CVPR}, pages 1--10, 2019.

\bibitem[Liu et~al.(2021)Liu, Lin, Cao, Hu, Wei, Zhang, Lin, and Guo]{liu2021swin}
Ze Liu, Yutong Lin, Yue Cao, Han Hu, Yixuan Wei, Zheng Zhang, Stephen Lin, and Baining Guo.
\newblock Swin transformer: Hierarchical vision transformer using shifted windows.
\newblock In \emph{ICCV}, pages 10012--10022, 2021.

\bibitem[Maas et~al.(2013)Maas, Hannun, Ng, et~al.]{maas2013rectifier}
Andrew~L Maas, Awni~Y Hannun, Andrew~Y Ng, et~al.
\newblock Rectifier nonlinearities improve neural network acoustic models.
\newblock In \emph{Proc. icml}, page~3. Atlanta, Georgia, USA, 2013.

\bibitem[Mangla et~al.(2020)Mangla, Kumari, Sinha, Singh, Krishnamurthy, and Balasubramanian]{mangla2020charting}
Puneet Mangla, Nupur Kumari, Abhishek Sinha, Mayank Singh, Balaji Krishnamurthy, and Vineeth~N Balasubramanian.
\newblock Charting the right manifold: Manifold mixup for few-shot learning.
\newblock In \emph{IEEE Win. Conf. Appl. Comput. Vis.}, pages 2218--2227, 2020.

\bibitem[Miller(1995)]{miller1995wordnet}
George~A Miller.
\newblock Wordnet: a lexical database for english.
\newblock \emph{Commun. ACM}, 38\penalty0 (11):\penalty0 39--41, 1995.

\bibitem[Mirza and Osindero(2014)]{mirza2014conditional}
Mehdi Mirza and Simon Osindero.
\newblock Conditional generative adversarial nets.
\newblock \emph{arXiv preprint arXiv:1411.1784}, 2014.

\bibitem[Mishra et~al.(2017)Mishra, Rohaninejad, Chen, and Abbeel]{mishra2017simple}
Nikhil Mishra, Mostafa Rohaninejad, Xi Chen, and Pieter Abbeel.
\newblock A simple neural attentive meta-learner.
\newblock \emph{arXiv preprint arXiv:1707.03141}, 2017.

\bibitem[OpenAI(2023)]{ChatGPT}
OpenAI.
\newblock Chatgpt.
\newblock \url{https://chat.openai.com}, 2023.

\bibitem[Oreshkin et~al.(2018)Oreshkin, Rodr{\'\i}guez~L{\'o}pez, and Lacoste]{oreshkin2018tadam}
Boris Oreshkin, Pau Rodr{\'\i}guez~L{\'o}pez, and Alexandre Lacoste.
\newblock Tadam: Task dependent adaptive metric for improved few-shot learning.
\newblock In \emph{NeurIPS}, 2018.

\bibitem[Peng et~al.(2019)Peng, Li, Zhang, Li, Qi, and Tang]{peng2019few}
Zhimao Peng, Zechao Li, Junge Zhang, Yan Li, Guo-Jun Qi, and Jinhui Tang.
\newblock Few-shot image recognition with knowledge transfer.
\newblock In \emph{ICCV}, pages 441--449, 2019.

\bibitem[Radford et~al.(2021)Radford, Kim, Hallacy, Ramesh, Goh, Agarwal, Sastry, Askell, Mishkin, Clark, et~al.]{radford2021learning}
Alec Radford, Jong~Wook Kim, Chris Hallacy, Aditya Ramesh, Gabriel Goh, Sandhini Agarwal, Girish Sastry, Amanda Askell, Pamela Mishkin, Jack Clark, et~al.
\newblock Learning transferable visual models from natural language supervision.
\newblock In \emph{Int. Conf. Mach. Learn.}, pages 8748--8763. PMLR, 2021.

\bibitem[Ravi and Larochelle(2017)]{ravi2017optimization}
Sachin Ravi and Hugo Larochelle.
\newblock Optimization as a model for few-shot learning.
\newblock In \emph{ICLR}, 2017.

\bibitem[Ren et~al.(2018)Ren, Triantafillou, Ravi, Snell, Swersky, Tenenbaum, Larochelle, and Zemel]{ren2018meta}
Mengye Ren, Eleni Triantafillou, Sachin Ravi, Jake Snell, Kevin Swersky, Joshua~B Tenenbaum, Hugo Larochelle, and Richard~S Zemel.
\newblock Meta-learning for semi-supervised few-shot classification.
\newblock \emph{arXiv preprint arXiv:1803.00676}, 2018.

\bibitem[Rusu et~al.(2018)Rusu, Rao, Sygnowski, Vinyals, Pascanu, Osindero, and Hadsell]{rusu2018meta}
Andrei~A Rusu, Dushyant Rao, Jakub Sygnowski, Oriol Vinyals, Razvan Pascanu, Simon Osindero, and Raia Hadsell.
\newblock Meta-learning with latent embedding optimization.
\newblock \emph{arXiv preprint arXiv:1807.05960}, 2018.

\bibitem[Satorras and Estrach(2018)]{satorras2018few}
Victor~Garcia Satorras and Joan~Bruna Estrach.
\newblock Few-shot learning with graph neural networks.
\newblock In \emph{ICLR}, 2018.

\bibitem[Schonfeld et~al.(2019)Schonfeld, Ebrahimi, Sinha, Darrell, and Akata]{schonfeld2019generalized}
Edgar Schonfeld, Sayna Ebrahimi, Samarth Sinha, Trevor Darrell, and Zeynep Akata.
\newblock Generalized zero-and few-shot learning via aligned variational autoencoders.
\newblock In \emph{CVPR}, pages 8247--8255, 2019.

\bibitem[Siyuan~Sun(2023)]{metaAdaM}
Hongyang~Gao Siyuan~Sun.
\newblock Meta-adam: A meta-learned adaptive optimizer with momentum for few-shot learning.
\newblock In \emph{NeurIPS}, 2023.

\bibitem[Snell et~al.(2017)Snell, Swersky, and Zemel]{snell2017prototypical}
Jake Snell, Kevin Swersky, and Richard Zemel.
\newblock Prototypical networks for few-shot learning.
\newblock In \emph{NeurIPS}, 2017.

\bibitem[Sohn et~al.(2015)Sohn, Lee, and Yan]{sohn2015learning}
Kihyuk Sohn, Honglak Lee, and Xinchen Yan.
\newblock Learning structured output representation using deep conditional generative models.
\newblock In \emph{NeurIPS}, 2015.

\bibitem[Sung et~al.(2018)Sung, Yang, Zhang, Xiang, Torr, and Hospedales]{sung2018learning}
Flood Sung, Yongxin Yang, Li Zhang, Tao Xiang, Philip~HS Torr, and Timothy~M Hospedales.
\newblock Learning to compare: Relation network for few-shot learning.
\newblock In \emph{CVPR}, pages 1199--1208, 2018.

\bibitem[Tian et~al.(2020)Tian, Wang, Krishnan, Tenenbaum, and Isola]{tian2020rethinking}
Yonglong Tian, Yue Wang, Dilip Krishnan, Joshua~B Tenenbaum, and Phillip Isola.
\newblock Rethinking few-shot image classification: a good embedding is all you need?
\newblock In \emph{ECCV}, pages 266--282. Springer, 2020.

\bibitem[Tokmakov et~al.(2019)Tokmakov, Wang, and Hebert]{tokmakov2019learning}
Pavel Tokmakov, Yu-Xiong Wang, and Martial Hebert.
\newblock Learning compositional representations for few-shot recognition.
\newblock In \emph{ICCV}, pages 6372--6381, 2019.

\bibitem[Tseng et~al.(2020)Tseng, Lee, Huang, and Yang]{tseng2020cross}
Hung-Yu Tseng, Hsin-Ying Lee, Jia-Bin Huang, and Ming-Hsuan Yang.
\newblock Cross-domain few-shot classification via learned feature-wise transformation.
\newblock In \emph{ICLR}, 2020.

\bibitem[Van~der Maaten and Hinton(2008)]{van2008visualizing}
Laurens Van~der Maaten and Geoffrey Hinton.
\newblock Visualizing data using t-sne.
\newblock \emph{J Mach Learn Res}, 9\penalty0 (11), 2008.

\bibitem[Vinyals et~al.(2016)Vinyals, Blundell, Lillicrap, Wierstra, et~al.]{vinyals2016matching}
Oriol Vinyals, Charles Blundell, Timothy Lillicrap, Daan Wierstra, et~al.
\newblock Matching networks for one shot learning.
\newblock In \emph{NeurIPS}, 2016.

\bibitem[Wah et~al.(2011)Wah, Branson, Welinder, Perona, and Belongie]{wah2011caltech}
Catherine Wah, Steve Branson, Peter Welinder, Pietro Perona, and Serge Belongie.
\newblock The caltech-ucsd birds-200-2011 dataset.
\newblock 2011.

\bibitem[Wang and Deng(2021)]{ijcai2021-149}
Haoqing Wang and Zhi-Hong Deng.
\newblock Cross-domain few-shot classification via adversarial task augmentation.
\newblock In \emph{IJCAI}, pages 1075--1081. International Joint Conferences on Artificial Intelligence Organization, 2021.

\bibitem[Wang et~al.(2019)Wang, Yu, Wang, Darrell, and Gonzalez]{wang2019tafe}
Xin Wang, Fisher Yu, Ruth Wang, Trevor Darrell, and Joseph~E Gonzalez.
\newblock Tafe-net: Task-aware feature embeddings for low shot learning.
\newblock In \emph{CVPR}, pages 1831--1840, 2019.

\bibitem[Xing et~al.(2019)Xing, Rostamzadeh, Oreshkin, and O~Pinheiro]{xing2019adaptive}
Chen Xing, Negar Rostamzadeh, Boris Oreshkin, and Pedro~O O~Pinheiro.
\newblock Adaptive cross-modal few-shot learning.
\newblock In \emph{NeurIPS}, 2019.

\bibitem[Xu and Le(2022)]{xu2022generating}
Jingyi Xu and Hieu Le.
\newblock Generating representative samples for few-shot classification.
\newblock In \emph{CVPR}, pages 9003--9013, 2022.

\bibitem[Yan et~al.(2021)Yan, Bouraoui, Wang, Jameel, and Schockaert]{yan2021aligning}
Kun Yan, Zied Bouraoui, Ping Wang, Shoaib Jameel, and Steven Schockaert.
\newblock Aligning visual prototypes with bert embeddings for few-shot learning.
\newblock In \emph{ICMR}, pages 367--375, 2021.

\bibitem[Ye et~al.(2020)Ye, Hu, Zhan, and Sha]{ye2020few}
Han-Jia Ye, Hexiang Hu, De-Chuan Zhan, and Fei Sha.
\newblock Few-shot learning via embedding adaptation with set-to-set functions.
\newblock In \emph{CVPR}, pages 8808--8817, 2020.

\bibitem[Zhang et~al.(2021)Zhang, Li, Ye, Huang, and Zhang]{zhang2021prototype}
Baoquan Zhang, Xutao Li, Yunming Ye, Zhichao Huang, and Lisai Zhang.
\newblock Prototype completion with primitive knowledge for few-shot learning.
\newblock In \emph{CVPR}, pages 3754--3762, 2021.

\bibitem[Zhang et~al.(2022)Zhang, Cai, Lin, and Shen]{zhang2022deepemd}
Chi Zhang, Yujun Cai, Guosheng Lin, and Chunhua Shen.
\newblock Deepemd: Differentiable earth mover's distance for few-shot learning.
\newblock \emph{IEEE TPAMI}, 2022.

\bibitem[Zhou et~al.(2017)Zhou, Lapedriza, Khosla, Oliva, and Torralba]{zhou2017places}
Bolei Zhou, Agata Lapedriza, Aditya Khosla, Aude Oliva, and Antonio Torralba.
\newblock Places: A 10 million image database for scene recognition.
\newblock \emph{IEEE Transactions on Pattern Analysis and Machine Intelligence}, 40\penalty0 (6):\penalty0 1452--1464, 2017.

\bibitem[Zhou et~al.(2023)Zhou, Wang, Zhang, Wei, and Zhang]{zhou2023revisiting}
Fei Zhou, Peng Wang, Lei Zhang, Wei Wei, and Yanning Zhang.
\newblock Revisiting prototypical network for cross domain few-shot learning.
\newblock In \emph{CVPR}, pages 20061--20070, 2023.

\bibitem[Zintgraf et~al.(2019)Zintgraf, Shiarli, Kurin, Hofmann, and Whiteson]{zintgraf2019fast}
Luisa Zintgraf, Kyriacos Shiarli, Vitaly Kurin, Katja Hofmann, and Shimon Whiteson.
\newblock Fast context adaptation via meta-learning.
\newblock In \emph{Int. Conf. Mach. Learn.}, pages 7693--7702. PMLR, 2019.

\end{thebibliography}
}


\end{document}